\documentclass[11pt,a4paper]{article}

\usepackage[T1]{fontenc}
\usepackage[utf8]{inputenc}
\usepackage{newtxtext}
\usepackage{newtxmath}
\usepackage[numbers,square,sort&compress]{natbib}
\usepackage{foodinsect-preprint}
\usepackage{tikz}
\usetikzlibrary{arrows.meta,positioning,shapes.geometric,calc}
\usepackage{graphicx}
\graphicspath{{figures/}}
\usepackage{booktabs}
\usepackage{amsmath}
\usepackage{amsfonts}
\usepackage{nicefrac}
\usepackage{multirow}
\usepackage{float}
\usepackage{wrapfig}
\usepackage{courier}
\usepackage{ulem}
\usepackage{multicol}

\DeclareMathOperator*{\argmin}{arg\,min}

\title{Batch-Invariant Spectral Intelligence for Robust and Explainable Insect Authentication}

\author{
Majharulislam Babor\textsuperscript{1,*},
Giacomo Rossi\textsuperscript{2},
Annalisa Altavilla\textsuperscript{1},\\
Oliver Schl\"{u}ter\textsuperscript{2},
and Marina M.-C. H\"{o}hne\textsuperscript{1,3}\\[0.7em]
\small\textsuperscript{1}Data Science in Bioeconomy, Leibniz Institute for Agricultural Engineering and Bioeconomy (ATB), \\
Potsdam, Germany\\
\small\textsuperscript{2}Systems Process Engineering, ATB, Potsdam, Germany\\
\small\textsuperscript{3}University of Potsdam, Potsdam, Germany\\[0.5em]
\small\textsuperscript{*}Corresponding author: \href{mailto:MBabor@atb-potsdam.de}{MBabor@atb-potsdam.de}
}

\date{}

\begin{document}

\maketitle

\begin{abstract}
\noindent Edible insects offer an efficient source of alternative protein, requiring less land and water and emitting less greenhouse gas than conventional livestock. However, their successful integration into the food supply 
chain demands reliable species authentication to control allergen exposure, prevent adulteration, and meet regulatory standards. Near-infrared spectroscopy provides a rapid analytical tool, but its performance drops 
when applied to production batches unseen during training due to batch-to-batch variation in spectral 
measurements. We introduce the Batch-Invariant Spectral Network (BISN), an end-to-end framework that combines a
learnable preprocessing module, initialised with Savitzky\textendash{}Golay filtering, with
an unsupervised, entropy-regularised auxiliary branch connected through a
gradient reversal layer. The branch uses no batch labels and supplies opposing
entropy gradients to the auxiliary head and preprocessing module.
In contrast to Domain-Adversarial Neural 
Network, which enforces domain adaptation only after feature extraction, BISN applies auxiliary entropy regularisation before the sparse encoder learns species-specific features. Using 2,700 spectra from three species (\textit{Acheta domesticus}, 
\textit{Hermetia illucens}, and \textit{Tenebrio molitor}) collected across three independent production 
batches, BISN achieves a mean leave-one-batch-out accuracy of $0.93$ (standard deviation $0.04$), outperforming
the strongest baseline by four percentage points ( $p=0.002$). Further insights gained by using explainable AI confirm that model 
decisions consistently rely on the lipid and protein absorption regions across all folds, connecting predictive 
performance to known insect biochemistry.
BISN addresses both cross-batch robustness and biochemical interpretability for automated insect species authentication 
in a controlled benchtop setting. The source code and dataset are publicly available at \url{https://github.com/majharB/bisn}. 
\end{abstract}

NIR spectroscopy; Food authenticity; Traceability; Domain
adaptation

\begin{multicols}{2}

\section{Introduction}

Meeting the growing global demand for dietary protein while reducing the environmental
burden of food production remains a central challenge for sustainable food
systems~\cite{Psarianos2025FRI,lange2021edible}.
Edible insects offer a promising protein alternative, converting feed to biomass
with high efficiency and substantially lower land, water, and greenhouse gas
requirements than conventional livestock~\cite{rossi2025insect,madau2020insect}.
Three species are of particular industrial relevance: the house cricket
\textit{Acheta domesticus}, the black soldier fly \textit{Hermetia illucens}, and the
yellow mealworm \textit{Tenebrio molitor}, each approved for human or animal consumption in
multiple jurisdictions and already incorporated into commercial protein
ingredients~\cite{wildbacher2025authentication}.

As insect production scales from artisanal rearing to industrial processing, reliable 
analytical tools are required not only to verify raw-material identity but also to 
ensure traceability and process consistency across production runs.
\textit{H.~illucens} larvae 
contain high lipid concentrations (29--43\,\% dry weight) dominated by 
medium-chain fatty acids such as lauric acid~\cite{Lawal2022, Borrelli2021}. 
\textit{T.~molitor} offers a higher protein fraction (48--75\,\% dry weight)
with an amino acid profile comparable to conventional animal proteins \cite{Laroche2019,vanHuis2013}, whereas 
\textit{A.~domesticus} provides 55--70\,\% protein \cite{Rumpold2013} and may contain substances that trigger cross-reactive allergenic 
responses~\cite{wildbacher2025authentication}. Misidentification risks product 
adulteration, regulatory non-compliance, and allergen exposure. While authoritative methods such as DNA 
barcoding and Liquid Chromatography-Tandem Mass Spectrometry (LC-MS/MS) proteomics deliver high accuracy, their processing 
time (typically requiring hours-to-days), expensive laboratory infrastructure, and laborious sample preparation limit their use in inline automatic process control.
Therefore, rapid, automated insect-monitoring methods have expanded to include image recognition, 
hyperspectral imaging, and bioacoustic 
monitoring~\cite{Tan2024,chakrabarty2026insectai}. 

Among rapid analytical techniques, near-infrared (NIR) spectroscopy is particularly suited 
for industrial applications because it simultaneously probes overtone and 
combination bands of O--H, C--H, and N--H bonds, yielding compositional 
information on water, lipids, proteins, and 
carbohydrates~\cite{Singh2026, Simmler2025, qu2020advances,burns2007handbook}. These 
spectral signatures can support rapid species discrimination under controlled acquisition conditions.
Although 
NIR-based classification has been demonstrated under controlled laboratory 
conditions~\cite{CruzTirado2026FoodControl, food_nir_review_2025, 
CruzTirado2025, Tan2024, CruzTirado2023FoodControl}, reliable industrial 
deployment is constrained by unwanted spectral variation due to batch effects.
{
Complementary LC--MS, NMR, ambient mass spectrometry such as DART--HRMS, and TXRF analyses characterise molecular and elemental profiles in insects~\cite{Aiello2023,Tata2022}.
These profiles vary with diet, developmental stage, and processing, creating batch differences that may affect NIR model performance~\cite{Bogusz2024,Zacometti2026,Ojha2021}.
}

Spectral variation across production batches arises from instrument drift, 
environmental fluctuations, variable sample moisture, and handling 
inconsistencies~\cite{rinnan2009review,zhao2019pls}. In insect processing, 
this variability is further amplified by the physical and chemical modifications 
imposed by production treatments, including blanching, plasma-activated water (PAW) 
exposure, and ultrasound application to facilitate protein extraction, each of which perturbs spectral regions 
that overlap with species-discriminative features. 
Blanching at $70\,^\circ\mathrm{C}$ induces protein denaturation, pigment 
degradation, and moisture redistribution, which alter NIR absorption band 
shapes and intensities~\cite{azzollini2018effects}. PAW exposure may oxidise lipids and modify protein 
structure, shifting signals in the C--H and N--H overtone 
regions~\cite{Ni2024Ultrasonics}. Ultrasound generates cavitation-induced 
microstructural disruption, releasing intracellular water and lipids while 
altering tissue porosity~\cite{rinnan2009review,Ni2024Ultrasonics}. Because 
these interventions perturb spectral regions that overlap with 
species-discriminative features, notably the lipid C--H overtone 
(840--910\,nm), protein N--H combination (1000--1180\,nm), and O--H stretching 
(910--1000\,nm) bands~\cite{burns2007handbook,qu2020advances}, achieving 
batch-robust classification requires addressing interference that extends 
beyond conventional multi-class NIR challenges under fixed conditions.
Addressing this challenge requires preprocessing or domain adaptation strategies that can 
decouple species-discriminative spectral structure from production batch and treatment-induced variation.

Classical preprocessing methods, including Savitzky--Golay filtering, standard normal variate (SNV) normalisation, and multiplicative scatter correction, 
effectively attenuate fixed baseline artefacts.
In practice, preprocessing configurations are typically selected by exhaustive 
grid search over transformation pipelines coupled with downstream classifiers, 
yet the combination that minimises in-distribution error provides no guarantee 
of generalisation to spectra acquired under unseen batch conditions as static 
transformations cannot adapt to batch-specific structure.
Calibration-transfer techniques like 
PDS~\cite{wang1991multivariate} and diPLS~\cite{nikzad2018domain} modify 
spectra directly but rely on static or weakly parametric mappings optimised 
independently of the downstream classifier.
Methods that 
couple domain adaptation with post hoc explanation, such as 
ShapDA~\cite{Babor2026ShapDA}, improve transferability but inherit feature 
selection from source-domain models and do not jointly optimise a 
batch-regularised preprocessing operator.
Similarly, attention-based architectures 
such as TabNet~\cite{arik2021tabnet} enable instance-wise feature selection 
but lack mechanisms to suppress spectroscopic batch effects.
Advanced domain-adversarial frameworks such as 
DANN~\cite{ganin2016dann} enforce domain 
invariance after feature extraction. In DANN, gradient-reversal constraints invert the gradient signal from a domain
discriminator, discouraging domain-specific structure in the latent
space. By operating at this level, such methods leave the input spectral
structure intact. This allows batch-correlated artefacts to persist in
the encoder's initial layers, potentially limiting robustness when the
target batch is entirely unseen during training.

Even if the model robustness would be increased, interpretability
of model decisions remains an independent requirement for industrial food
authentication.
A model achieves practical utility only when it identifies which spectral 
regions encode species information, which are altered by processing 
variation, and why individual predictions fail. These requirements extend 
beyond scalar accuracy and demand a representation whose geometric structure 
can be mapped directly to species biochemistry.

To address these gaps, we propose the Batch-Invariant Spectral Network (BISN), 
a framework for robust and interpretable insect species classification across 
production batches. 
Our main contributions are threefold. 
(a) We introduce Batch-Invariant Spectral Network (BISN), an end-to-end architecture 
that is robust in insect identification using milled insect–water slurries measured with a benchtop NIR spectrometer. 
It applies unsupervised auxiliary entropy regularisation upstream of the
sparse encoder and species classifier. Unlike conventional DANN training~\cite{ganin2016dann},
the auxiliary branch does not use physical batch labels.
Here, upstream means that the entropy-regularised branch acts after the learnable convolution, but before the sparse encoder and species classifier.
(b) We characterise how blanching, plasma-activated water exposure, and 
ultrasound reshape NIR signatures across wavelength regions, 
establishing a spectroscopic and biochemical baseline for model 
interpretation.
(c) We bridge predictive performance and biochemical explanation through
Integrated Gradients attributions and region-constrained counterfactual
optimisation. This analysis identifies which spectral regions drive
species discrimination and which remain structurally uninformative in
the learned representation.
Independent commercial insect batches, factorial processing conditions, and strict held-out-batch evaluation provide a robust test of cross-batch generalisation. The publicly available code and data make this evidence transparent and reproducible.
\section{Materials and Methods}

\subsection{Experimental Design and Data Collection}
\label{sec:experimental_design}

 The study included three commercially relevant edible insect species approved for food or feed use: \textit{Acheta domesticus}, \textit{Hermetia illucens}, and
\textit{Tenebrio molitor}~\cite{rossi2025insect}.
Adult \textit{A.~domesticus} were sourced from Tropic-Shop (Nordhorn, Germany), fifth-stage larvae of \textit{H.~illucens} from Hermetia
Baruth GmbH (Baruth/Mark, Germany), and fifth-stage larvae of
\textit{T.~molitor} from Fauna Topics Zoobedarf Zucht- und Handels GmbH (Marbach am Neckar, Germany).
Each species involved three independent production batches. We purchased insects separately for each batch to ensure biological independence. Conducting measurements on separate days integrated acquisition effects with batch material variation. 
This design represents combined batch and acquisition variation that can occur across commercial production lots.

A full factorial design was employed to incorporate species-specific
biochemical variation from process-induced and measurement-related effects.
Within each batch, whole insects were mixed with water and milled using
a Grindomix GM200 (Retsch GmbH, Haan, Germany) at 10{,}000\,rpm for
three cycles of 30\,s each, yielding a homogeneous slurry. The
insect-to-water ratio was 1:1 for \textit{H.~illucens} and
\textit{T.~molitor}, and 1:2 for \textit{A.~domesticus}, as determined
by preliminary tests to obtain a fluid consistency suitable for
reliable ultrasound propagation. The additional water for \textit{A.~domesticus} is consistent with its protein-rich composition and the high water-holding capacity~\cite{LucasGonzalez2019}. Samples were then assigned to three
treatments. Treatment T0 served as the reference condition, using tap
water. Treatment T1 applied blanching at $70\,^{\circ}\mathrm{C}$ for
five minutes, with insects immersed in water prior to milling, inducing
protein denaturation, pigment degradation, and moisture
redistribution~\cite{azzollini2018effects,burns2007handbook}.
Treatment T2 replaced tap water with plasma-activated water (PAW), a
non-thermal process in which plasma-generated reactive oxygen and
nitrogen species modify lipid and protein residues and induce partial
pigment bleaching, shifting NIR responses~\cite{Ni2024IJBM}.

Each treatment group was subdivided into two ultrasound conditions: U0
(no ultrasound) and U1 (ultrasound applied). Ultrasound treatment was
performed using a UIP1000hdT sonicator (Hielscher Ultrasonics GmbH,
Teltow, Germany) equipped with a D4-1.8 booster and a CS4d40L2
sonotrode (diameter 40\,mm, amplitude factor 1.2). Processing
conditions were: volume 700\,mL, power 700\,W (power density
${\approx}1\,\mathrm{W\,mL^{-1}}$), frequency 20\,kHz, acoustic
intensity $6\,\mathrm{W\,cm^{-2}}$, amplitude 100\%, and total
treatment time 30\,min with a pulsed cycle of 10\,s on and 50\,s off
(effective sonication time 5\,min, cycle ${\approx}15\,\%$).
Temperature was maintained below $50\,^{\circ}\mathrm{C}$ using an ice
jacket monitored by a thermocouple. The acoustic cavitation caused by
ultrasound disrupts cellular membranes, releases intracellular water
and lipids, and alters tissue porosity, modifying the scattering
coefficient and effective NIR signals through physical rather than
compositional mechanisms~\cite{Ni2024IJBM}.

The design ($3\,\text{batches} \times 3\,\text{species} \times
3\,\text{treatments} \times 2\,\text{ultrasound conditions}$) produced
$2{,}700$ labelled spectra, with $50$ spectra per factorial cell, yielding marginal totals of $n = 900$ per
species, batch, and treatment, and $n = 1{,}350$ per ultrasound condition (Supplementary~\ref{sec:dataset}).
Each batch corresponds to a distinct commercial purchase, acquired at
30--60 day intervals (November 2024, January 2025, and February 2025)
to capture both biological and day-specific spectral variation.

NIR spectra were acquired in reflectance mode using a PerkinElmer
Lambda 950 UV-Vis/NIR spectrometer (PerkinElmer, Waltham, MA, USA)
controlled by UV WinLab software. The light source was a tungsten lamp,
and detection employed a PbS/PMT combination with a detector changeover
at 860.80\,nm. Spectra were recorded with an integration time of 1\,s per cycle, a detector
slit width of 2\,nm, and source and detector attenuators both set to
10\,\%.
Samples were measured in front-face mode by placing a thin layer
($2\,\mathrm{mm}$) in a custom circular sample holder
(diameter 12\,mm). The instrument was calibrated against a
WS-1-SL white reflectance standard (Labsphere Inc., North Sutton, NH,
USA) prior to each measurement session and recalibrated every 6\,h.
All spectra were collected at room temperature ($20 \pm 2\,^{\circ}
\mathrm{C}$) and controlled relative humidity ($50 \pm 5\,\%$).

Each spectrum was acquired from an independently drawn sample within each batch, treatment, and ultrasound condition. Spectra were
recorded over $700$--$2050\,\mathrm{nm}$ at $10\,\mathrm{nm}$ intervals,
yielding 136 wavelength variables. This window encompasses the principal
overtone and combination bands associated with water, lipids, and
proteins~\cite{burns2007handbook,food_nir_review_2025,swir_review_2015,
tsenkova2009aquaphotomics}, the dominant chemical constituents of insect
biomass. The recorded spectral signal is reported throughout as intensity in arbitrary units (a.u.).

\subsection{Classical Preprocessing for Spectra}
\label{sec:preprocessing}

Preprocessing
methods are widely used to suppress measurement artefacts while preserving biochemically meaningful
variation, particularly across batches where differences in instrument state,
environmental conditions, and sample handling accumulate. In this work, each production batch is treated as an independent domain. To prevent
information leakage, all preprocessing parameters were optimised exclusively
on the training batches.

The pipeline considered several standard spectral transformations. Standard
normal variate (SNV) normalisation and multiplicative scatter correction
(MSC)~\cite{barnes1989standard} address scattering
effects from differences in particle size and packing density but differ in whether correction
parameters are estimated from a reference mean spectrum (MSC) or from each
spectrum independently (SNV)~\cite{rinnan2009review}. Polynomial detrending
was evaluated to compensate for baseline
shifts~\cite{rinnan2009review}. Savitzky--Golay (SG)
filtering~\cite{savitzky1964smoothing} was applied for noise reduction with first- or second-derivative transformation to resolve overlapping
features and suppress baseline shifts~\cite{rinnan2009review}.
The hyperparameter search space comprised the following five categories: normalisation from
\{none, SNV, MSC\}, detrending degree from \{0, 1, 2\}, SG window size from
odd integers in \{11, 15, 21, 25, 31, 35, 41, 45, 51, 55, 61, 65, 71, 75\},
SG polynomial order from \{1, 2, 3\}, and SG derivative order from \{0, 1, 2\}. This classical preprocessing search was conducted independently for each LOBO fold.

For each LOBO fold, configurations were evaluated via five-fold stratified
cross-validation on the training batches. The performance was measured by two metrics: the balanced accuracy of a
logistic regression classifier discriminating the two training batches, and the
Maximum Mean Discrepancy (MMD)~\cite{gretton2012kernel} between batch spectral
distributions using an RBF kernel with bandwidth set by the median heuristic.
Both metrics were minimised, as lower values indicate reduced batch
separability and closer distributional alignment, respectively.

Configurations were ranked across both criteria using a rank-sum rule rather
than direct metric averaging. Because batch-discrimination accuracy and MMD
operate on different numerical scales with different dispersion characteristics,
rank aggregation provides a scale-invariant, robust selection criterion that
prevents either objective from dominating due to its variance or magnitude.
The configuration with the lowest mean rank across inner folds was selected
for each LOBO split.

The selected configuration was then refitted on the full training batches and
applied to the held-out test batch prior to model training, ensuring complete
independence from test data. Species separability was deliberately excluded
as an optimisation criterion to keep preprocessing model-agnostic and avoid
favouring any particular downstream learner. To verify that the strategy does
not suppress species-relevant variation, exploratory analyses were conducted
on three spectral representations: raw spectra, classically preprocessed
spectra, and the learned representation produced by the BISN preprocessing
module.

\subsection{Exploratory Data Analysis}
\label{sec:eda}

To characterise dominant sources of spectral variation and evaluate the
impact of preprocessing, exploratory analyses were conducted on three
representations: raw spectra, classically preprocessed spectra, and representation by the BISN preprocessing module. The goal is to quantify variation attributable to
insect species relative to technical and process-induced variation from
production batch, treatment (T0, T1, T2), and ultrasound exposure.

\vspace{0.3cm}
\noindent\textit{(a) Multivariate analysis of variance}\\
Principal component analysis (PCA) was applied to spectral representations to
obtain a low-dimensional structure. Since PCA does not
formally test whether experimental factors (e.g., batch) influence joint spectral structure,
multivariate analysis of variance (MANOVA) was performed on the first five
principal components, which together explained $\geq 90\,\%$ of variance across representations. Wilks' $\Lambda$ and the corresponding $F$-statistic were
computed per factor to assess whether group centroid means differed
significantly in the PC subspace. Wilks' $\Lambda$ ranges from
0 to 1. A lower value indicates a stronger multivariate effect of the factor
relative to the residual. Restricting
MANOVA to five PCs excludes variance in higher-order components.
Therefore, we used Silhouette and batch-probe metrics on the full representation as complementary measures for the cluster separation and batch mixing tests.

\vspace{0.3cm}
\noindent\textit{(b) Cluster separation and batch mixing}\\
Silhouette coefficients~\cite{roussseeuw1987silhouettes} were computed for
both species and batch labels. Higher species silhouette indicates stronger
biological separation. The batch silhouette values near zero or negative indicate
that batch identity does not correspond to geometrically compact clusters.
The batch-probe tests whether batch identity remains linearly recoverable from the representations.
Batch classification was assessed by fitting a logistic regression model on a representation using five-fold cross-validation stratified jointly by
batch and species. The classifier
was trained on four folds and evaluated on the remaining fold. The results are
reported as mean $\pm$ standard deviation across folds.

\subsection{Machine Learning Models}

\subsubsection{Batch-Invariant Spectral Network (BISN)}
\label{sec:bisn}

\begin{figure*}[t]
\centering
\includegraphics[width=\textwidth]{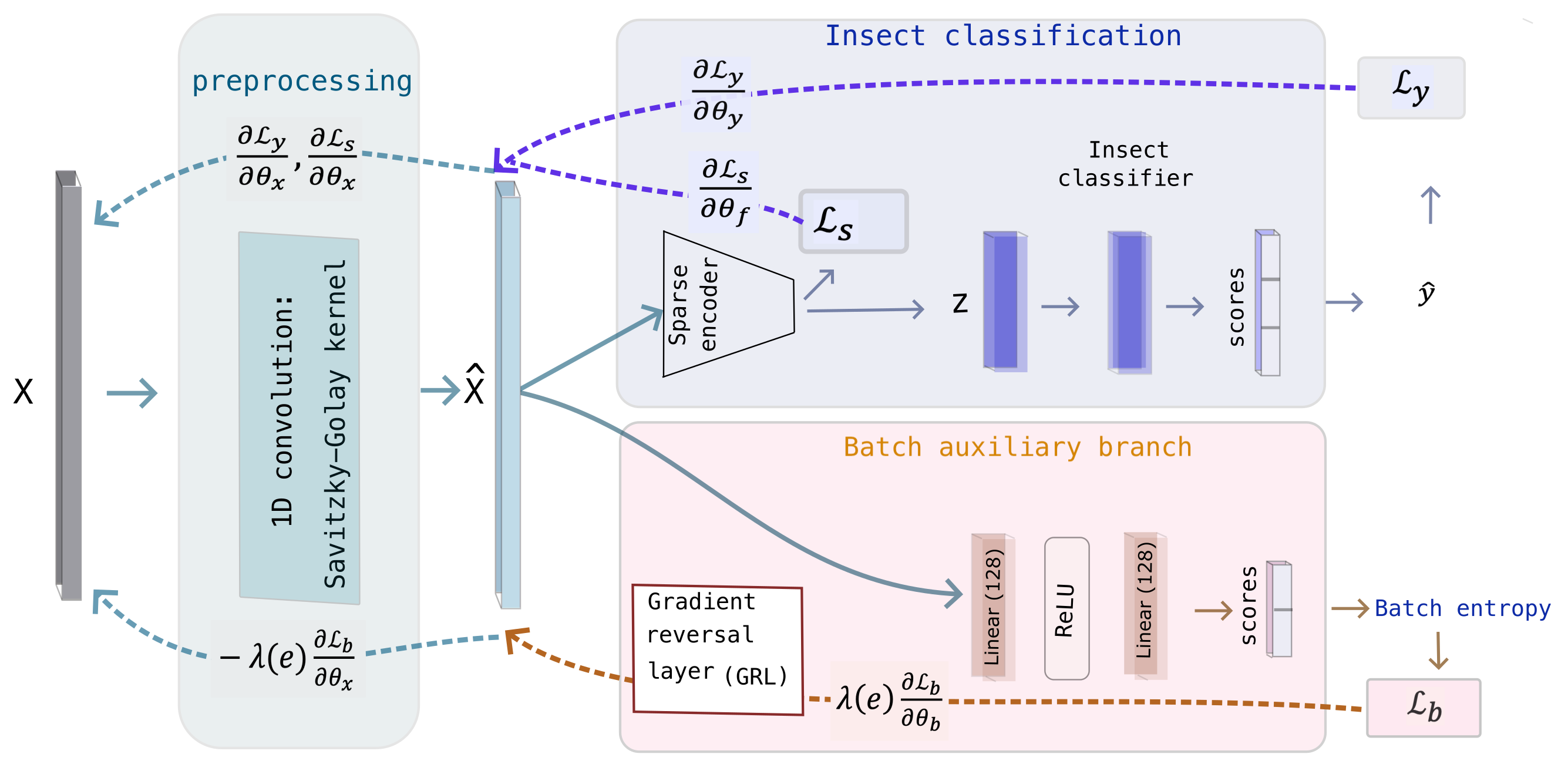}
\caption{Architecture of the Batch-Invariant Spectral Network (BISN).
Raw NIR spectra $x$ enter an informed preprocessing module consisting of a
Savitzky--Golay-initialised learnable 1D convolution followed by instance
normalisation, producing a learned preprocessing output $\mathbf{\hat{x}}$.
This representation feeds into a sparse attentive encoder that generates a compact
latent embedding $\mathbf{z}$ for insect species classification.
Species logits are computed through a linear classifier and optimised via the species
cross-entropy loss $\mathcal{L}_y$.
In parallel, $\hat{\mathbf{x}}$ is passed through a gradient reversal
layer (GRL) to a two-layer batch auxiliary head. The auxiliary head is trained by minimising the negative-entropy loss
$\mathcal{L}_{b}$, which increases the entropy of its output distribution.
The GRL reverses this gradient before it reaches the preprocessing module.
Thus, the auxiliary head and preprocessing module are optimised in opposing
entropy directions. The auxiliary outputs are unsupervised slots and do not
represent physical batch classes.
The sparse-mask entropy regularisation $\mathcal{L}_\mathrm{s}$ is computed from the
sparse attention mask. It regularises both the encoder and the preprocessing module through the composite gradient update.
Solid arrows denote the forward pass and dotted arrows indicate the backward gradient flow.
The three-way interaction of $\mathcal{L}_y$, $\mathcal{L}_b$, and
$\mathcal{L}_\mathrm{s}$ implements a joint objective that preserves
species-relevant biochemical information while suppressing batch-dependent variation.}
\label{fig:bisn_architecture}
\end{figure*}

Each spectrum $\mathbf{x} \in \mathbb{R}^d$ ($d = 136$) is paired
with a species label $y \in \{1,2,3\}$ and a production-batch label
$b \in \{1,2,3\}$.
The goal is to learn a preprocessed spectral representation
$\mathbf{\hat{x}} = g(\mathbf{x};\theta_x)$ 
that preserves species-discriminative biochemical structure while reducing production batch associated variation, i.e.,
$$
P(\mathbf{\hat{x}}\mid y)
\approx
P(\mathbf{\hat{x}}\mid y,b).
$$
BISN achieves this with four jointly optimised components
(Fig.~\ref{fig:bisn_architecture}):
(1)~a learnable, Savitzky--Golay-initialised preprocessing module that maps
$\mathbf{x}\in \mathbb{R}^{136}$ to a batch-regularised preprocessing output $\mathbf{\hat{x}}\in \mathbb{R}^{136}$,
(2)~a sparse attentive encoder that maps $\mathbf{\hat{x}}$ to a latent embedding
$\mathbf{z} \in \mathbb{R}^{8}$,
(3)~a linear species classifier, and
(4)~a batch auxiliary entropy branch connected to the preprocessing module
via a gradient reversal layer (GRL).
The joint training objective is
\begin{equation}
 \mathcal{L}
 = \mathcal{L}_y + \beta\,\mathcal{L}_s + \lambda(e)\,\mathcal{L}_b,
 \label{eq:joint_loss}
\end{equation}
where $\mathcal{L}_y$ is the species cross-entropy, 
$\mathcal{L}_s \leq 0$ is the sparse-mask entropy regularisation (Component~2),
and
$\mathcal{L}_b \leq 0$ is the batch auxiliary entropy loss (Component~4), $\beta > 0$ is a scalar weight controlling the
contribution of the sparse-mask entropy regularisation term, and
the batch auxiliary entropy contribution is annealed as $\lambda(e) \in [0,1]$.
The GRL negates the gradient of $\mathcal{L}_b$ before it reaches the preprocessing module, such that the auxiliary head and preprocessing module receive opposing entropy
gradients.
It reduces batch-identifying structure from $\mathbf{\hat{x}}$.
Each component is described in detail below.

\vspace{0.3cm}
\noindent\textit{Component 1: Informed Preprocessing Module}\\
This module is designed to reduce batch-induced spectral variation through a
Savitzky--Golay (SG)-initialised 1D convolution followed by per-spectrum
instance normalisation.
The convolution kernel is initialised with the analytic first-derivative
SG coefficients $\{a^{(1)}_j\}$:
\begin{equation}
 W_j \leftarrow a^{(1)}_j,
 \qquad
 \tilde{\mathbf{x}} = \mathbf{W} \ast \mathbf{x},
 \qquad \mathbf{W} \in \mathbb{R}^{1 \times w}.
 \label{eq:conv}
\end{equation}
The SG specific parameters, i.e., fold-optimal window $w$,
polynomial degree $r$, and degree of derivative $k$ (described in Section ~\ref{sec:preprocessing}) were identified by the classical preprocessing grid
search and offer a physically grounded initialisation of the convolutional kernel based on
established spectroscopic data preprocessing. Rather than imposing a fixed analytical filter, BISN treats this initialisation as an informed starting point that embeds prior knowledge about noise suppression, baseline correction, and resolution of overlapping absorption bands directly into the optimisation process.
By jointly training the kernel through the species-classification objective and the entropy-regularised 
 batch auxiliary branch, BISN learns a spectral representation that selectively suppresses dataset-specific batch artefacts while preserving the biochemical structure required for robust species discrimination across unseen production batches.
The convolution output is normalised per instance, such that sample-specific intensity offsets are removed and global spectral scalings are stabilised independently of the
batch composition during training, preventing batch-dependent amplitude variation from propagating into the learned representation ($\mathbf{\hat{x}}$).

\vspace{0.3cm}
\noindent\textit{Component 2: Sparse Attentive Encoder}\\
Starting from the preprocessed
spectrum $\mathbf{\hat{x}}$, a feature transformer $F_{\mathrm{split}}$ first produces an intermediate representation $\mathbf{a} \in \mathbb{R}^{n_a}$ that encodes
spectral patterns relevant for wavelength selection. 
This representation is linearly projected into the spectral dimension, stabilised via Ghost Batch Normalisation (GBN), and transformed through sparsemax \cite{martins2016}
into a sparse wavelength-attention mask:
\begin{equation}
    \begin{aligned}
        \mathbf{M}
        &= \mathrm{sparsemax}\,\!\bigl(\mathrm{GBN}\,(W_a\,\mathbf{a})\bigr)
        \in \Delta^{d-1}, \\
        &\quad W_a \in \mathbb{R}^{d \times n_a}.
    \end{aligned}
    \label{eq:sparsemax_mask}
\end{equation}
Sparsemax projects scores onto the nearest point of the probability simplex, setting weights to exactly zero for wavelengths below
an adaptive threshold (Supplementary~\ref{app:sparsemax}).
The mask is then applied element-wise to the preprocessed and normalised (GBN) spectrum: $\mathbf{x}_{\mathrm{sel}}=\mathbf{M}\otimes\bar{\mathbf{x}}$, yielding a spectrally filtered representation that retains only the most biochemically informative wavelength regions for downstream classification.
The masked spectrum
is then processed by a step transformer $F_{\mathrm{step}}$ that shares its
first two layers with $F_{\mathrm{split}}$ but extends the representation through additional refinement layers to generate the final embedding
$\mathbf{z} \in \mathbb{R}^{8}$.
This shared-weight design couples wavelength selection and representation learning, ensuring that the sparse mask is optimised jointly with the downstream species-discriminative embedding rather than as an isolated feature-selection stage. To prevent the attention mask from collapsing onto a single wavelength, we introduce an
entropy-based regularisation term:
\begin{equation}
 \mathcal{L}_s
 = \frac{1}{N}\sum_{n=1}^{N}\sum_{i=1}^{d}
 M_{i}^{(n)}\log\,\,\!\bigl(M_{i}^{(n)} + \epsilon\bigr),
 \label{eq:sparsity_loss}
\end{equation}
where $M_i^{(n)} \in [0,1]$ denotes the mask weight assigned to wavelength $i$ from sample $n$, and $\epsilon$ is a small constant added for numerical
stability. Minimising $\mathcal{L}_s$ counteracts an excessive concentration of attention and distributes importance across a compact set of complementary wavelengths, thereby aligning the learned representation with broader biochemical absorption structure rather than with isolated spectral peaks.

\vspace{0.3cm}
\noindent\textit{Component 3: Species Classifier}\\
The species classifier maps the latent embedding $\mathbf{z} \in \mathbb{R}^{8}$ to class scores using a single linear layer
followed by a softmax function:
\begin{equation}
    \begin{aligned}
        \hat{\mathbf{y}}
        &= \mathrm{softmax}\,\,(W_y\,\mathbf{z} + \mathbf{b}_y), \\
        &\quad W_y \in \mathbb{R}^{3 \times 8},
        \quad \mathbf{b}_y \in \mathbb{R}^{3}.
    \end{aligned}
    \label{eq:species_softmax}
\end{equation}
Afterward, the classifier is trained by minimising the cross-entropy loss:
\begin{equation}
 \mathcal{L}_y
 = -\frac{1}{N}\sum_{n=1}^{N}\sum_{c=1}^{3}
 y_{n,c}\log\,\,\!\left(\hat{y}_{n,c} + \epsilon\right),
 \label{eq:species_loss}
\end{equation}
where $y_{n,c}$ is the one-hot encoded ground-truth label for sample $n$
and class $c$, $\hat{y}_{n,c}$ is the corresponding predicted probability. A linear
classification head ensures that all non-linear
discriminative capacities are learned within the feature representation $\mathbf{z}$.

\vspace{0.3cm}
\noindent\textit{Component 4: Batch Auxiliary Entropy Branch}\\
BISN includes an unsupervised auxiliary branch acting on the preprocessed
representation $\mathbf{\hat{x}}$.
 The branch operates on the preprocessed representation $\mathbf{\hat{x}}$ through a Gradient Reversal Layer (GRL), which leaves the forward pass unchanged but reverses the gradient sign during backpropagation. 
A two-layer auxiliary head produces the logits: 
\begin{equation}
 \begin{aligned}
 \mathbf{b}
 &= W_2\,\mathrm{ReLU}(W_1\mathbf{\hat{x}} + \mathbf{b}_1) + \mathbf{b}_2, \\
 W_1 &\in \mathbb{R}^{128\times d},\quad
 W_2 \in \mathbb{R}^{B' \times 128}, \\
 \mathbf{b}_1 &\in \mathbb{R}^{128},
 \end{aligned}
 \label{eq:discriminator}
\end{equation}
where $B'=2$ is the number of auxiliary output slots. These slots do not
correspond to the identities of the two source batches. The auxiliary
probabilities are:
$\mathbf{q}^{(n)} =
\mathrm{softmax}(\mathbf{b}^{(n)}) \in \Delta^{B'-1}$,
where $\mathbf{q}^{(n)}$ denotes the auxiliary probability distribution
produced for spectrum $n$.
 The batch auxiliary entropy loss is:
\begin{equation}
 \mathcal{L}_b
 = \frac{1}{N}\sum_{n=1}^{N}\sum_{\ell=1}^{B'}
 q_{\ell}^{(n)}\log\, \,\!\bigl(q_{\ell}^{(n)} + \epsilon\bigr),
 \label{eq:negentropy}
\end{equation}
where $\epsilon$ is a small numerical constant for stability. Minimising $\mathcal{L}_b$ with respect to the auxiliary-head
parameters maximises the entropy of its output distribution. The GRL reverses
this gradient for the preprocessing parameters, which therefore receive
$-\lambda(e)\nabla\mathcal{L}_{b}$. The two modules are consequently
optimised in opposing auxiliary-entropy directions. This objective supplies
an unsupervised adversarial regularisation signal but does not explicitly
encode physical batch identity.

\subsubsection{Baseline Classification Models}
\label{sec:baseline_models}

We evaluated a range of classification models spanning classical
statistical methods, chemometric approaches, domain-adaptation
techniques, and recent deep learning architectures (Table \ref{tab:baseline_models}), alongside our
proposed Batch-Invariant Spectral Network (BISN). All models were trained and evaluated under a consistent leave-one-batch-out
(LOBO) protocol. For each fold, hyperparameters were selected by
five-fold stratified cross-validation on the training batches using
species accuracy, after which the chosen configuration was trained using the internal training--validation split and evaluated on the held-out batch.

For all baseline models, inputs consisted of raw and classically preprocessed
spectra (Section~\ref{sec:preprocessing}), with preprocessing parameters
optimised exclusively on the training batches to avoid information
leakage. In contrast, BISN operates directly on raw spectra via its
learnable preprocessing module.

\begin{table*}[t]
\centering
\caption{Summary of baseline models grouped by methodological class}
\begin{tabular}{lll}
\toprule
\textbf{Model class} & \textbf{Method} & \textbf{Key characteristic} \\
\midrule
Linear / probabilistic
& LDA~\cite{fisher1936use}
& Linear separability via variance maximisation \\
& GPC~\cite{rasmussen2006gaussian}
& Bayesian non-parametric classification (RBF kernel) \\
\midrule
Chemometric
& PLSDA~\cite{barker2003partial}
& Latent projection maximising covariance with labels \\
& diPLS~\cite{nikzad2018domain}
& Domain-adaptive PLS with automatic component selection \\
& PDS-PLSDA~\cite{wang1991multivariate}
& Piecewise spectral alignment across batches \\
\midrule
Domain adaptation
& ShapDA~\cite{Babor2026ShapDA}
& Domain-invariant representation with interpretability \\
& DANN~\cite{ganin2016dann}
& Adversarial feature alignment via gradient reversal \\
\midrule
Deep learning
& SpectraTr~\cite{fu2022spectratr}
& Transformer with spectral self-attention \\
& NIRCoreVision-MLP~\cite{singh2025nircorevision}
& CNN-based feature extraction with core-set selection \\
& TabPFN~\cite{hollmann2025accurate}
& Prior-data-fitted transformer for tabular inference \\
& TabNet~\cite{arik2021tabnet}
& Sparse sequential attention over features \\
\bottomrule
\end{tabular}
\label{tab:baseline_models}
\end{table*}

Classical methods (LDA, GPC) provide transparent baselines for linear and
kernel-based separability. Chemometric approaches (PLSDA, diPLS,
PDS-PLSDA) explicitly model latent spectral structure and include
calibration-transfer mechanisms for handling inter-batch variability.
Domain-adaptation methods (ShapDA, DANN) aim to suppress batch-specific
information in the learned representation. Deep learning architectures (SpectraTr, NIRCoreVision-MLP,
TabPFN, TabNet) capture non-linear spectral dependencies through
attention or hierarchical feature extraction.
Detailed hyperparameter search spaces for all models are provided in
Supplementary (Table~\ref{tab:all_hyperparameters}b).

\subsubsection{Performance Evaluation}
\label{sec:metrics}
Classification performance is reported as mean $\pm$ standard deviation
across LOBO folds.
Accuracy is defined as
\begin{equation}
\text{Acc} = \frac{1}{n} \sum_{i=1}^{n} \mathbf{1}\,(\hat{y}_i = y_i),
\label{eq_acc}
\end{equation}
where $n$ is the number of test samples, $y_i$ is the true species
label of sample $i$, $\hat{y}_i$ is the predicted label, and
$\mathbf{1}(\cdot)$ is the indicator function equal to 1 if the
condition holds and 0 otherwise.
The F1 score across classes is computed as follows:
\begin{equation}
\text{F1} = \frac{1}{C} \sum_{c=1}^{C}
\frac{2\,\text{TP}_c}{2\,\text{TP}_c + \text{FP}_c + \text{FN}_c},
\label{eq_caf}
\end{equation}
where $C$ is the number of classes, and
$\text{TP}_c$, $\text{FP}_c$, $\text{FN}_c$ are the true positives,
false positives, and false negatives for class $c$, respectively.
Model generalisation was evaluated using the LOBO
protocol, in which each of the three production batches was held out
once as a strictly independent test set.
The remaining two batches ($n = 1{,}800$ spectra) constituted the
training pool, where samples were further divided into a training
subset ($80\,\%$) and an internal validation subset
($20\,\%$) using a split stratified by class.
The internal validation subset was used exclusively for early stopping and checkpoint selection, while final performance was assessed only on the held-out test batch.
This restriction also applied to DANN, diPLS, ShapDA, and PDS-PLSDA, where adaptation and calibration used only the two training batches. For DANN, these batches served as the internal source and target domains, while PDS-PLSDA used no transfer spectra from the held-out batch.

\section{Results and Discussion}

\subsection{Spectral Shifts Across Species
and Treatments}
\label{sec:spectral_shifts}

\begin{figure*}[t]
\centering
\includegraphics[width=\textwidth]{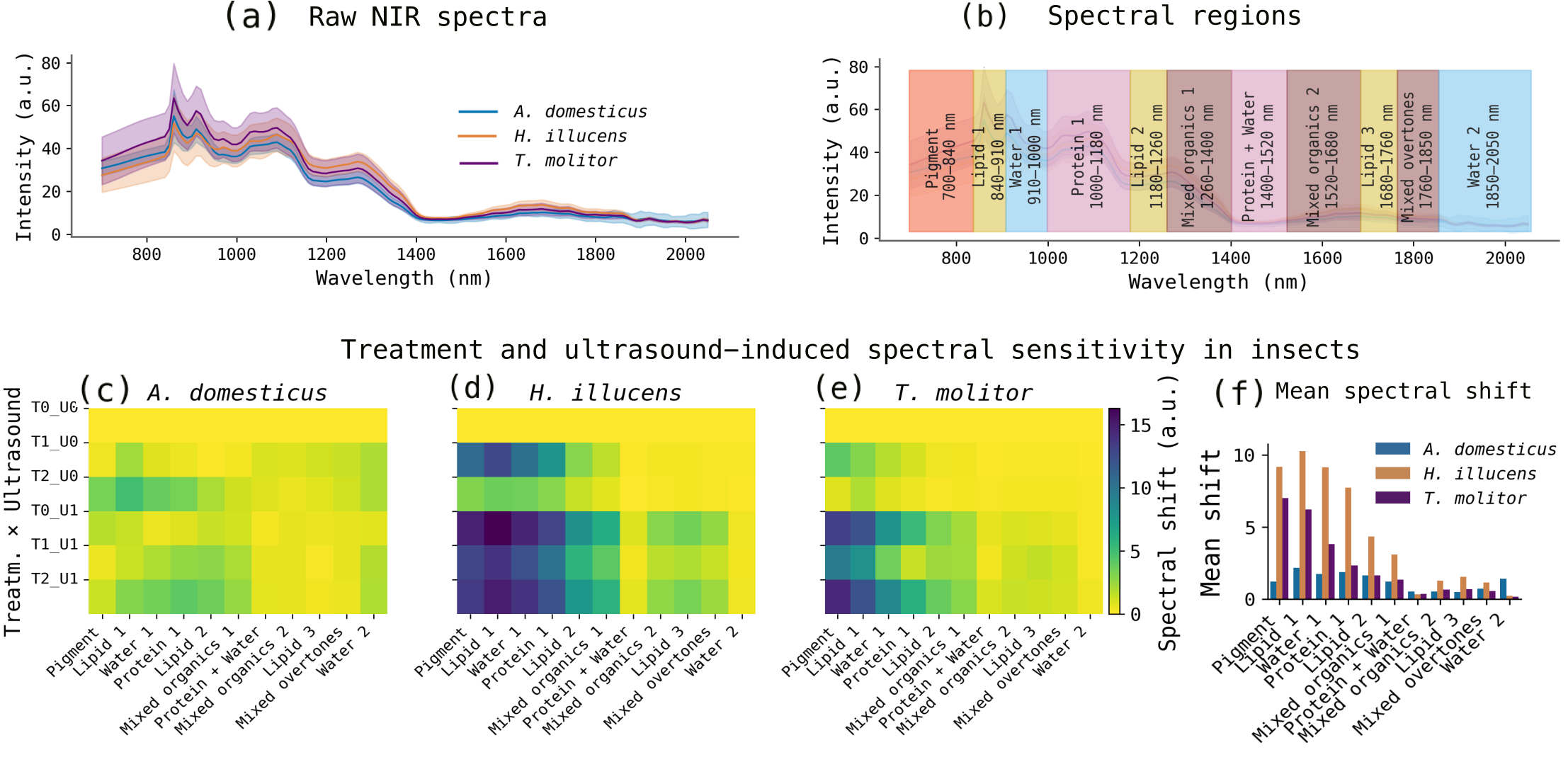}
\caption{Spectral sensitivity analysis across insect species, processing treatments,
and chemically defined NIR wavelength regions.
\textbf{(a)} Mean raw NIR spectra showing species-specific
baseline profiles and wavelength-dependent variability.
\textbf{(b)} The shaded bands indicate the eleven chemically defined regions detailed in
Supplementary~\ref{app:wavelength_regions}.
\textbf{(c--e)} Region-level spectral perturbation heatmaps for insect species.
The rows correspond to the treatment--ultrasound combination relative to the control
(T0\_U0: fresh insects in tap water, no ultrasound), and the columns refer to the eleven different spectral regions.
Colour intensity encodes the absolute mean spectral deviation from
the control.
\textit{H.~illucens} exhibits the largest and most widespread perturbations,
concentrated in pigment, lipid~1, water~1, and protein~1 under ultrasound-assisted
PAW (T2\_U1). \textbf{(f)} Mean spectral shift per region aggregated across treatments and
ultrasound conditions, revealing species-specific susceptibility profiles.}
\label{fig:spectral_shift}
\end{figure*}

We analyse how spectral variability is
distributed across species, processing conditions, and wavelength regions.
From the raw mean spectra, shown in Fig.~\ref{fig:spectral_shift}a, we can observe consistent
species-specific spectral intensity profiles, which reflect biochemical composition differences between species. \textit{T.~molitor} exhibits the highest overall intensity. Substantial within-species
variability indicates that
batch, treatment, and ultrasound effects introduce considerable
spectral noise that could obscure biological signals.

To better understand where these treatment-induced changes occur and to relate them to underlying chemical components, we divided the NIR spectrum into eleven established wavelength regions associated with
pigments, lipids, water, proteins, and mixed overtone signals (Fig.~\ref{fig:spectral_shift}b, Supplementary~\ref{app:wavelength_regions}).
Spectral changes were quantified as the absolute change in mean spectral intensity relative to the reference condition (T0\_U0: corresponding to fresh insects in tap water, no ultrasound). Values are reported in arbitrary units (a.u.), representing the magnitude of spectral change rather than a physical concentration.
The regional shifts presented in
Figs.~\ref{fig:spectral_shift}c--e reveal a consistent pattern across species.
\textit{H.~illucens} exhibits the strongest perturbations across
most wavelength regions, \textit{T.~molitor} shows moderate perturbations,
and \textit{A.~domesticus} remains the least affected.
Across all three species, the pigment, lipid~1, water~1, protein~1,
lipid~2, and mixed-organics~1
regions consistently show the largest mean deviations, while
water~2, lipid~3, mixed-organics~2, protein--water, and
mixed-overtone regions remain comparatively stable.

Blanching (T1\_U0) produces the largest perturbations among non-ultrasound conditions, while PAW without ultrasound (T2\_U0) induces only minor deviations across all species. Ultrasound substantially amplifies perturbations under all treatment conditions, with the PAW combination (T2\_U1) producing the largest shifts observed.
These findings have direct implications for modelling.

The spectral regions most affected by treatment- and ultrasound-induced perturbations---pigment, lipid~1, water~1, protein~1, lipid~2, and mixed-organics~1---are not randomly distributed. These regions overlap with the same spectral bands that carry the strongest biological signal for species discrimination~\cite{rinnan2009review,
food_nir_review_2025}.
A robust classifier must preserve species-discriminative information while suppressing the batch- and process-induced distortions. 

\subsection{Variance Decomposition and Representation Analysis}
\label{sec:representation_analysis}
To assess whether preprocessing reorganises the balance between species- and batch-related variance, we examined three representations: raw spectra, classical preprocessed spectra, and the representation of the BISN preprocessing component. The classical preprocessing pipeline for folds~1 and~2 (MSC,
second-order polynomial detrend as baseline removal, SG first derivative, $w=61$,
$r=1$) was applied to all spectra.
The optimal preprocessing configurations for three folds are summarised in
Supplementary~\ref{sec:best_classical_preprocessing}.

\begin{figure*}[t]
\centering
\includegraphics[width=\textwidth]{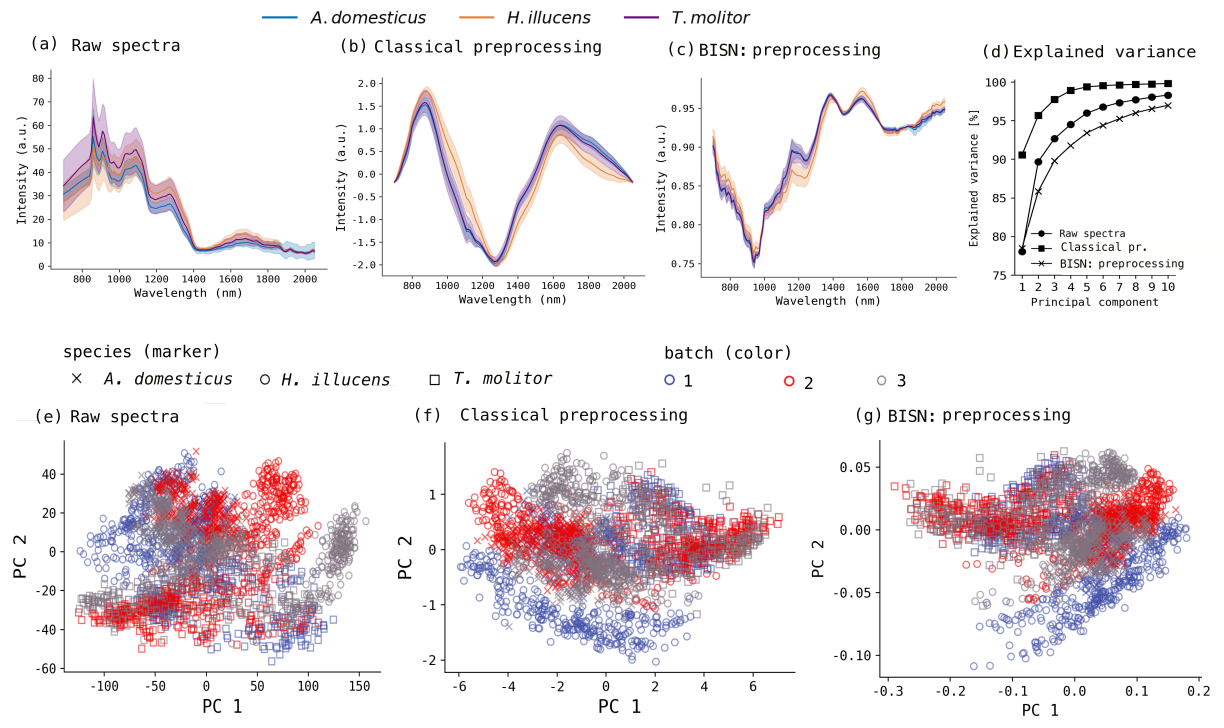}
\caption{
NIR spectral profiles across batches, species, and preprocessing stages.
\textbf{Top row}: mean spectra coloured by species identity
(\textit{A.~domesticus}: blue, \textit{H.~illucens}: orange, \textit{T.~molitor}:
purple) for \textbf{(a)} raw, \textbf{(b)} classically preprocessed, \textbf{(c)}
BISN preprocessing output, and \textbf{(d)} Explained variance ratio for the first ten principal components across
the three representations, showing that classical preprocessing concentrates variance
in PC1 while the BISN representation distributes it more evenly across components.
\textbf{Bottom row}: PCA-based comparison of spectral representations across batches and species.
\textbf{(e--g)} PCA scatter plots of (e) raw NIR spectra, (f) classically preprocessed
spectra, and (g) BISN preprocessing output from component 1.
Each point represents a sample spectrum, where marker shape indicates species identity
(\textit{A.~domesticus}: cross, \textit{H.~illucens}: circle,
\textit{T.~molitor}: square) and colour indicates production batch
(batch 1: blue, batch 2: red, batch 3: grey).
}
\label{fig:spectra_pca}
\end{figure*}

Fig.~\ref{fig:spectra_pca} presents the spectral representations and their principal component projections.
Classical preprocessing concentrates variance in PC1 due to SG
first-derivative amplification, while the BISN preprocessing output distributes it
more evenly across components (Fig.~\ref{fig:spectra_pca}d), indicating a broader encoding of species-relevant structure. In the raw PCA scatter (Fig.~\ref{fig:spectra_pca}e), batch-dependent stratification dominates, with \textit{T.~molitor} and \textit{H.~illucens} forming diffuse, partially overlapping distributions.
Classical preprocessing reduces within-species scatter but leaves residual
batch offsets, especially for \textit{T.~molitor} (Fig.~\ref{fig:spectra_pca}f).
In contrast, the BISN preprocessing output 
(Fig.~\ref{fig:spectra_pca}g) yields compact, well-separated species
clusters with data from batches distributed uniformly within each species
cloud.

\begin{figure*}[t]
\centering
\includegraphics[width=\linewidth]{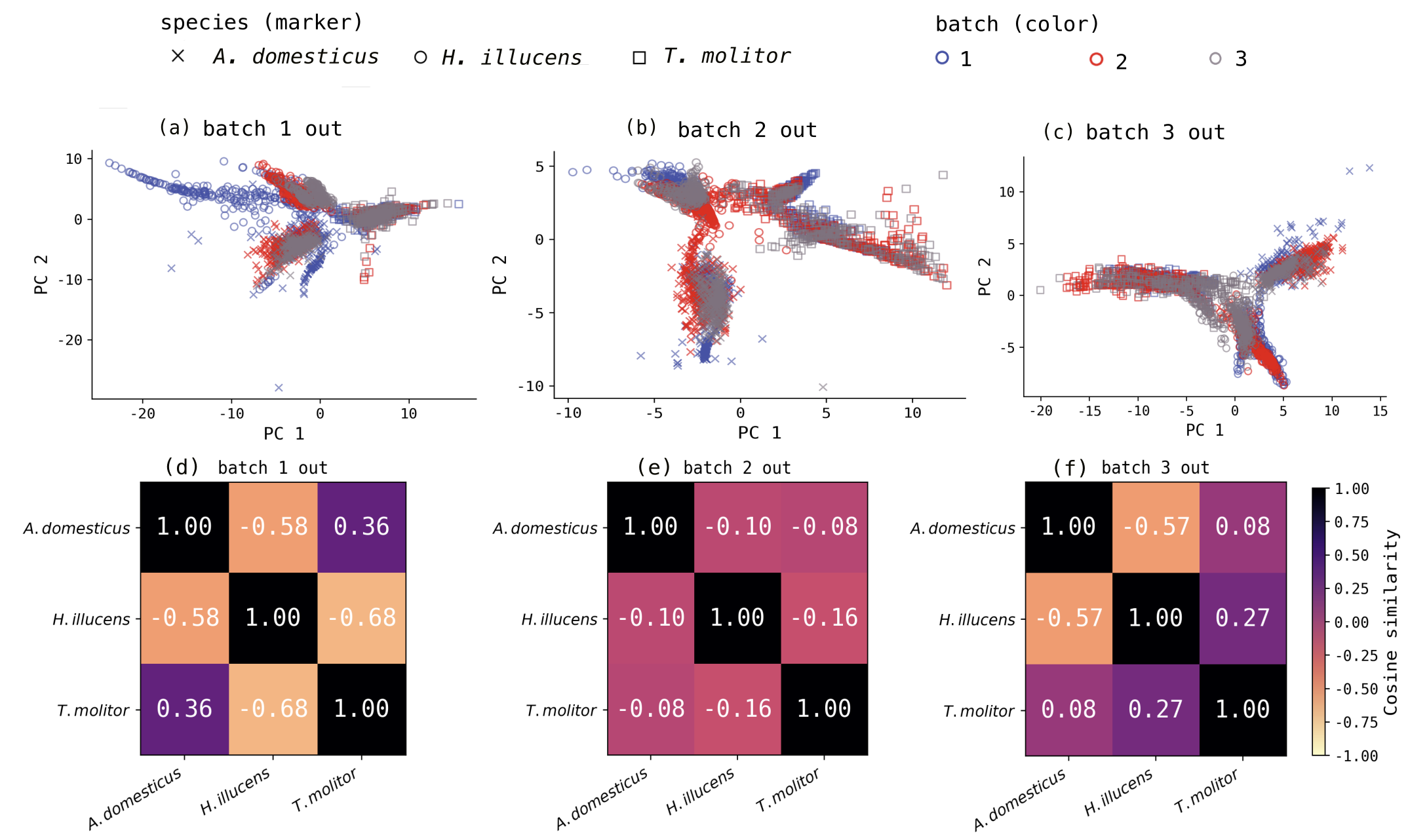}
\caption{BISN latent embedding visualisation across leave-one-batch-out (LOBO) folds. The signed latent embedding was obtained after the encoder’s final linear projection, rather than from its intermediate ReLU-activated features.
\textbf{(a--c)} PCA projections of the 8-dimensional BISN embedding for
LOBO folds.
Marker shapes denote batch identity and colours denote species. (d--f) Cosine similarity matrices of mean BISN embeddings across insect
species for LOBO folds 1, 2, and 3.
AD: \textit{A.~domesticus}; HI: \textit{H.~illucens}; TM: \textit{T.~molitor}.
Off-diagonal values reflect inter-species
embedding similarity.
Fold~1 exhibits the highest inter-species similarity between
\textit{A.~domesticus} and \textit{T.~molitor} ($0.36$); fold~3 shows
the closest proximity between \textit{H.~illucens} and
\textit{T.~molitor} ($0.27$).}
\label{fig:bisn-embedding}
\end{figure*}

Table~\ref{tab:latent-metrics} quantifies these observations. Raw spectra
yield a species silhouette of $0.07$ and batch-probe accuracy of
$0.95 \pm 0.01$, confirming that samples are more easily separated by batch than by species. The raw batch silhouette of $-0.03$ indicates that batch
clusters are not compact, consistent with stratified rather than clustered batch structure.
Classical preprocessing reduces batch-probe accuracy to
$0.43$--$0.44$ across folds, but species silhouette remains low
($0.08$--$0.20$) and inconsistent. The BISN preprocessing output achieves
comparable species silhouette ($0.10$--$0.14$) but higher batch-probe
accuracy in folds 1 and 3 ($0.75 \pm 0.02$ and $0.71 \pm 0.02$),
indicating fold-dependent residual batch content at this intermediate
stage. The BISN embedding yields the clearest restructuring with species
silhouette rising to $0.56$--$0.60$ across all folds, a four- to six-fold
improvement over raw spectra and classical preprocessing. The batch
silhouette remains near zero and batch-probe accuracy falls to
$0.50$--$0.55$, confirming that the sparse attentive encoder suppresses
residual batch structure not removed by the preprocessing module alone. However, it remains above the three-class chance level accuracy of 0.33.

\begin{table*}[t]
\centering
\caption{Quantitative analysis of insect class structure and batch dependence within leave-one-batch-out (LOBO) protocol across representations.
Silhouette scores ($\mathrm{Sil}$) are reported for species and batch labels. Batch Acc.\ denotes the accuracy of the three-class batch probe. Chance accuracy is 0.33 and values closer to 0.33 indicate lower linear batch recoverability.}
\begin{tabular}{lcccccc}
\toprule
\textbf{Representation} & \textbf{LOBO} & $\mathbf{Sil_\mathrm{species}}$ & $\mathbf{Sil_\mathrm{batch}}$ & \textbf{Batch Acc.} \\
\midrule
Raw spectra & $-$ & $0.07$ & $-0.03$ & $0.95 \pm 0.01$ \\
\midrule
\multirow{2}{*}{Classical preproc.} & $1$,$2$ & $0.08$ & $-0.03$ & $0.43 \pm 0.02$ \\
 & $3$ & $0.20$ & $-0.02$ & $0.44 \pm 0.01$ \\
\midrule
\multirow{3}{*}{BISN preproc. output ($\hat{X}$)}
 & 1 & $0.13$ & $-0.00$ & $0.75 \pm 0.02$ \\
 & 2 & $0.10$ & $-0.01$ & $0.44 \pm 0.02$ \\
 & 3 & $0.14$ & $-0.00$ & $0.71 \pm 0.02$ \\
 \midrule
\multirow{3}{*}{BISN embedding ($Z$)}
 & 1 & $0.56$ & $-0.04$ & $0.50 \pm 0.01$ \\
 & 2 & $0.60$ & $-0.00$ & $0.53 \pm 0.02$ \\
 & 3 & $0.56$ & $-0.03$ & $0.55 \pm 0.01$ \\
\bottomrule
\end{tabular}
\label{tab:latent-metrics}
\end{table*}

MANOVA on the first five principal components
(Table~\ref{tab:manova_cmp}) confirms these findings. Raw spectra show
species as the dominant factor ($\Lambda = 0.25 \pm 0.00$,
$F = 528 \pm 8$), but with a species-to-batch $F$ ratio of only
$12 \pm 6$. Classical preprocessing reduces the treatment effect
($F$: $34 \pm 7$ to $12 \pm 4$) but weakens species discriminability
($F$: $471 \pm 103$) and amplifies the ultrasound effect
($F$: $219 \pm 10$ to $379 \pm 38$), reflecting first-derivative 
amplification of sharp ultrasound-induced spectral contrasts. The BISN
preprocessing output maintains species discriminability
($F = 515 \pm 60$) while the batch $F$ rises to $106 \pm 9$, reflecting
concentration of residual batch variance into five PCs. In the
BISN embedding, although the batch effect remained comparable to raw spectra, the species effect increased sixfold
($\Lambda = 0.02 \pm 0.01$, $F = 3194 \pm 963$) and the species-to-batch
$F$ ratio rises to $85 \pm 72$, far exceeding all other representations. Large standard deviations reflect fold-to-fold variability.

\begin{table*}[t]
\centering
\caption{MANOVA results on five principal components ($\geq\,90\,\%$ variance). The principal component analysis was fitted on the training batches and applied to the combined train--test data ($n=2,700$) within each LOBO fold. Reported values are mean $\pm$ standard deviation across the three LOBO folds. Lower Wilks' $\Lambda$ and higher $F$ indicate a stronger multivariate effect of the corresponding factor.}
\label{tab:manova_cmp}
\resizebox{\textwidth}{!}{%
\begin{tabular}{lccccccccc}
\toprule
& \multicolumn{2}{c}{\textbf{Species}} 
& \multicolumn{2}{c}{\textbf{Batch}} 
& \multicolumn{2}{c}{\textbf{Treatment}} 
& \multicolumn{2}{c}{\textbf{Ultrasound}} 
& \textbf{Species/Batch} \\
\cmidrule(lr){2-3}\cmidrule(lr){4-5}\cmidrule(lr){6-7}\cmidrule(lr){8-9}
\textbf{Representation}
& $\Lambda$ & $F$
& $\Lambda$ & $F$
& $\Lambda$ & $F$
& $\Lambda$ & $F$
& $F$ ratio \\
\midrule
Raw spectra
& $0.25 \pm 0.00$ & $528 \pm 8$
& $0.82 \pm 0.06$ & $58 \pm 22$
& $0.88 \pm 0.02$ & $34 \pm 7$
& $0.71 \pm 0.01$ & $219 \pm 10$
& $12 \pm 6$ \\

Classical preprocessing
& $0.29 \pm 0.05$ & $471 \pm 103$
& $0.82 \pm 0.02$ & $56 \pm 8$
& $0.96 \pm 0.01$ & $12 \pm 4$
& $0.59 \pm 0.03$ & $379 \pm 38$
& $9 \pm 3$ \\

BISN preproc. output ($\hat{X}$)
& $0.26 \pm 0.03$ & $515 \pm 60$
& $0.70 \pm 0.02$ & $106 \pm 9$
& $0.87 \pm 0.01$ & $40 \pm 4$
& $0.78 \pm 0.03$ & $152 \pm 26$
& $5 \pm 0$ \\

BISN embedding ($Z$)
& $0.02 \pm 0.01$ & $3194 \pm 963$
& $0.82 \pm 0.07$ & $57 \pm 24$
& $0.92 \pm 0.02$ & $22 \pm 6$
& $0.77 \pm 0.12$ & $182 \pm 125$
& $85 \pm 72$ \\
\bottomrule
\end{tabular}%
}
\end{table*}

The BISN latent embedding (Fig.~\ref{fig:bisn-embedding}) shows compact,
well-separated species clusters across all LOBO folds, with samples from
different batches distributed homogeneously within each species cloud.
Cosine similarity matrices confirm this structure (Fig.~\ref{fig:bisn-embedding}d--f). The inter-species
similarities in fold~2 fall within $[-0.16, -0.08]$, indicating
well-separated and partially opposing class centroids. Fold~1 shows the largest positive similarity between \textit{T.~molitor} and \textit{A.~domesticus}
of $0.36$, reflecting closer centroid and
elevated mutual confusion between them
(Fig.~\ref{fig:bisn-embedding}a, d). Fold~3 shows a positive
inter-species similarity between \textit{T.~molitor} and
\textit{H.~illucens} ($0.27$), suggesting partial centroid alignment and a potential source of confusion between these two species.

\subsection{Cross-Batch Classification Performance}
\label{sec:classification}
In the following, we examine the performance of various models, which is summarised in Table~\ref{tab:insect_classification}.
On raw spectra, baselines achieve accuracies of $0.66$--$0.81$, with high
fold-to-fold variability driven by sensitivity to batch-specific spectral
offsets. DANN is the strongest baseline on raw spectra ($0.81 \pm 0.15$),
yet its large standard deviation indicates inconsistent generalisation
across held-out batches. Classical preprocessing improves all baselines
substantially, with the largest gains for non-adversarial methods. TabNet
rises from $0.75$ to $0.88$ ($+0.13$) and TabPFN from $0.74$ to $0.87$
($+0.13$). Classical calibration transfer methods do not exceed $0.77$ after
preprocessing, remaining below simpler classifiers such as GPC ($0.85$).
DANN gains only $+0.08$ ($0.81 \rightarrow 0.89$) from classical preprocessing,
compared to TabNet, TabPFN and GPC.
This suggests that DANN's adversarial training partially compensates for batch-induced 
spectral variation, leaving less room for improvement from classical preprocessing.

In contrast, BISN achieves $0.93 \pm 0.04$ in both accuracy and macro F1, the highest
values across all models. Compared with TabNet on raw spectra, the gain of $18$ percentage
points ($0.93$ vs. $0.75$) is directly attributable to the learnable
preprocessing and batch auxiliary branch. The improvement over the strongest
baseline, DANN on preprocessed spectra ($0.89 \pm 0.05$), is four
percentage points in both metrics, with fold-to-fold standard deviation
reduced from $\pm 0.05$ to $\pm 0.04$. Bootstrap resampling over $10^6$ iterations
confirmed that this difference is statistically significant, with an
accuracy difference of $0.041$, 95\,\% CI $[0.023, 0.067]$, and
$p < 10^{-6}$ (Supplementary~\ref{app:bootstrap_comparison}a).
A complementary paired preparation-block bootstrap produced an equal-weight mean difference of \(0.041\) (95\,\% CI: \(0.013\)–\(0.074\); \(p=0.002\); Supplementary Table~\ref{tab:bootstrap_comparison}b), conditional on the three observed production batches.

\begin{table*}[t]
\centering
\small
\caption{Cross-batch classification performance under leave-one-batch-out (LOBO)
evaluation.
Results are mean $\pm$ standard deviation across three LOBO folds.
BISN operates directly on raw spectra via its internal learnable preprocessing module
and is therefore not evaluated on separately preprocessed spectra.
The best baseline result (DANN, preprocessed) is typeset in bold for reference.}
\label{tab:insect_classification}
\begin{tabular}{lcccc}
\toprule
& \multicolumn{2}{c}{\textbf{Raw spectra}} &
\multicolumn{2}{c}{\textbf{Preprocessed spectra}} \\
\cmidrule(lr){2-3}\cmidrule(lr){4-5}
\textbf{Model} & {Acc} & {F1 score} & {Acc} & {F1 score} \\
\midrule
LDA & $0.75 \pm 0.09$ & $0.75 \pm 0.08$ & $0.75 \pm 0.08$ & $0.74 \pm 0.08$ \\
GPC & $0.78 \pm 0.12$ & $0.78 \pm 0.12$ & $0.85 \pm 0.06$ & $0.85 \pm 0.06$ \\
diPLS & $0.67 \pm 0.15$ & $0.67 \pm 0.15$ & $0.77 \pm 0.06$ & $0.77 \pm 0.06$ \\
PLSDA & $0.71 \pm 0.09$ & $0.70 \pm 0.09$ & $0.74 \pm 0.09$ & $0.74 \pm 0.09$ \\
PDS-PLSDA & $0.71 \pm 0.08$ & $0.70 \pm 0.08$ & $0.77 \pm 0.09$ & $0.77 \pm 0.09$ \\
ShapDA & $0.69 \pm 0.10$ & $0.66 \pm 0.10$ & $0.77 \pm 0.10$ & $0.74 \pm 0.11$ \\
SpectraTr & $0.66 \pm 0.14$ & $0.64 \pm 0.15$ & $0.71 \pm 0.04$ & $0.69 \pm 0.04$ \\
NIRCoreVision-MLP & $0.75 \pm 0.08$ & $0.73 \pm 0.11$ & $0.86 \pm 0.05$ & $0.86 \pm 0.06$ \\
TabPFN & $0.74 \pm 0.10$ & $0.73 \pm 0.07$ & $0.87 \pm 0.08$ & $0.86 \pm 0.08$ \\
TabNet & $0.75 \pm 0.09$ & $0.75 \pm 0.09$ & $0.88 \pm 0.06$ & $0.88 \pm 0.06$ \\
DANN & $0.81 \pm 0.15$ & $0.81 \pm 0.15$ & $\mathbf{0.89 \pm 0.05}$ & $\mathbf{0.89 \pm 0.05}$ \\ 
\midrule
\textbf{BISN} & $\mathbf{0.93 \pm 0.04}$ & $\mathbf{0.93 \pm 0.04}$ & $-$ & $-$ \\
\bottomrule
\end{tabular}
\end{table*}

\begin{figure*}[t]
\centering
\includegraphics[width=\linewidth]{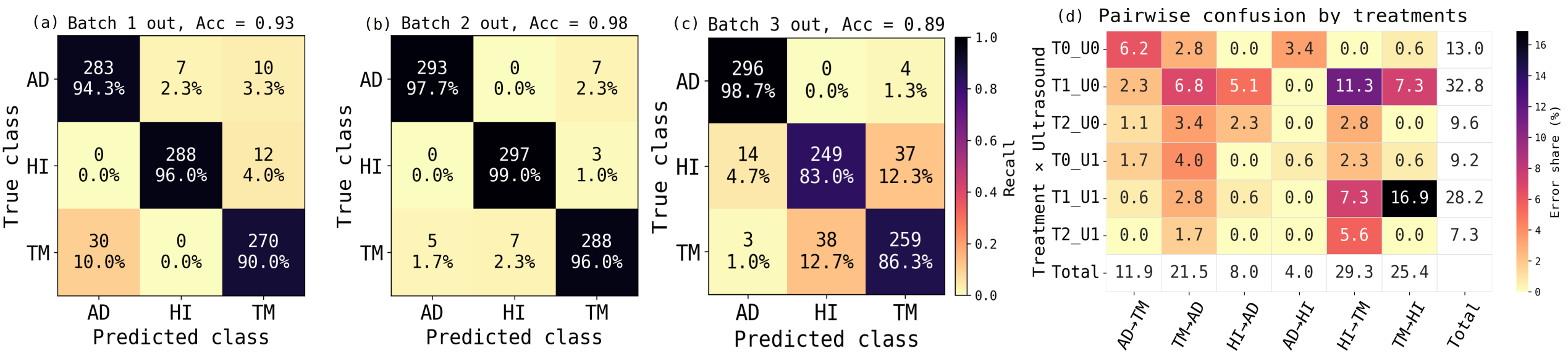}
\caption{Confusion matrices for BISN under leave-one-batch-out evaluation.
\textbf{(a)} Batch~1 held out; \textbf{(b)} Batch~2 held out; \textbf{(c)} Batch~3
held out.
Rows are true species labels, columns are predicted labels.
AD: \textit{A.~domesticus}; HI: \textit{H.~illucens}; TM: \textit{T.~molitor}.
Colour intensity is the recall score.
Fold~2 achieves near-perfect classification; fold~3 shows elevated confusion between
\textit{H.~illucens} and \textit{T.~molitor}. \textbf{(d)} Misclassification stratified by treatments (T0: tap water, T1: blanching, T2: PAW---plasma activated water) and ultrasound condition (U0: not applied, U1: applied)
across all three LOBO folds.
TM--HI confusion is concentrated in T1 (T1\_U0, T1\_U1), validating the positive cosine similarity in embedding space between TM--HI in Fig.~\ref{fig:bisn-embedding}f. The PAW treatment is associated with modest confusion.}
\label{fig:conf_matrix}
\end{figure*}

Per-fold confusion patterns are presented in Fig.~\ref{fig:conf_matrix}.
Fold~2 achieves near-perfect classification with only 22 misclassifications
out of 900 test samples, per-species recalls of $97.7\,\%$, $99.0\,\%$,
and $96.0\,\%$ for \textit{A.~domesticus}, \textit{H.~illucens}, and
\textit{T.~molitor}. The results are consistent with the narrow cosine similarity range in the fold~2 embedding ( Fig.~\ref{fig:bisn-embedding}e). Fold~1 yields 59 misclassifications, dominated by $10\,\%$ of \textit{T.~molitor} samples
predicted as \textit{A.~domesticus}, reflecting their partial centroid
alignment in BISN embeddings ( Fig.~\ref{fig:bisn-embedding}a,~\ref{fig:bisn-embedding}d) and highest cosine similarity ($0.36$). Fold~3 is the most challenging, with
96 misclassifications, concentrated along the
\textit{H.~illucens}$\rightarrow$\textit{T.~molitor} confusion axis, where recalls drop to
$83.0\,\%$ and $86.3\,\%$. This pattern is attributable to the cosine similarity of
$0.27$ between their latent embeddings
( Fig.~\ref{fig:bisn-embedding}c,~\ref{fig:bisn-embedding}f).

Among the 177 total misclassifications, blanching contributes $61.0\,\%$
of errors and PAW the least ($16.9\,\%$) as shown in Fig.~\ref{fig:conf_matrix}d. The
\textit{H.~illucens}$\rightarrow$\textit{T.~molitor} axis accounts for $54.7\,\%$
of all errors, driven by blanching-induced thermal denaturation and
moisture redistribution, which reduce the compositional contrast between
these two species. Under blanching with ultrasound (T1\_U1),
\textit{T.~molitor}\,$\rightarrow$\,\textit{H.~illucens} errors rise
from $7.3\,\%$ to $16.9\,\%$ while
\textit{H.~illucens}\,$\rightarrow$\,\textit{T.~molitor} decreases from
$11.3\,\%$ to $7.3\,\%$, an asymmetry consistent with cavitation-driven
intracellular lipid release shifting \textit{T.~molitor} spectra toward
the \textit{H.~illucens} space~\cite{Ni2024IJBM}.
\textit{A.~domesticus} remains well-classified across all conditions,
with residual errors attributable to fold-specific centroid proximity.

Several studies have employed spectroscopic and image-based machine learning approaches for edible insect authentication, providing a benchmark for the results presented here. Ni et al. applied NIR spectroscopy (800--2500 nm) combined with PLS regression and PCA to detect adulteration in insect protein powders among the same three species investigated in the present work, achieving RMSECV values ranging from 1.8 to 4.1\,\% \cite{Ni_2024_FoodControl}. Although these prediction errors are low, the study was conducted on a dataset of 311 samples acquired under controlled conditions from a single source and time point. Cruz-Tirado et al. evaluated FTIR and two portable NIR spectrometers for edible insect flour authentication using 301 samples, reporting 100\,\% sensitivity and specificity for \textit{Alphitobius diaperinus} and \textit{Tenebrio molitor} authentication using FTIR, and class efficiency (CEFF) values of 93--100\,\% with the 1450--2450 nm spectrometer \cite{CruzTirado2026FoodControl}. Despite these strong classification results, the study did not address compositional variability arising from biological batch differences or inter-supplier variation. Zhang et al. demonstrated that SVM applied to ATR-FTIR spectra of empty fly puparia achieved 100\,\% accuracy within the biological fingerprint region (1800--1300 cm$^{-1}$) across five fly species on a dataset of 450 spectra \cite{Zhang2024SAA}. However, the study was conducted under highly controlled forensic conditions using specimens from a single rearing facility, limiting its applicability to multi-source food processing scenarios. Ma et al. proposed a PCA-VGG architecture integrating hyperspectral spatial-spectral features (400--1000 nm) for the classification of three morphologically similar cricket species from the genus \textit{Teleogryllus}. The study attained an overall accuracy of 88\,\% via stratified 5-fold cross-validation with transfer learning on a dataset of only 48 specimens \cite{Ma2025OptLaserTech}. While each of the aforementioned studies demonstrates strong performance within its respective experimental scope, all were conducted on small to moderate datasets under homogeneous acquisition conditions. In contrast, our study employs 2,700 spectra spanning three temporally distinct batch acquisitions, covering the most widely applied insect processing treatments. Under these demanding generalisation conditions, the proposed BISN model achieved a classification accuracy of 0.93, demonstrating competitive performance relative to the cited works while extending the evaluation to temporally distinct production batches under controlled benchtop acquisition.

\subsection{Interpretability of BISN}
\label{sec:explain_bisn}
To interpret the learned representations, we analysed spectral attributions via Integrated Gradients and decision boundaries through spectral region-constrained counterfactual optimisation.

\begin{figure*}[!htbp]
\centering
\includegraphics[width=0.9\linewidth]{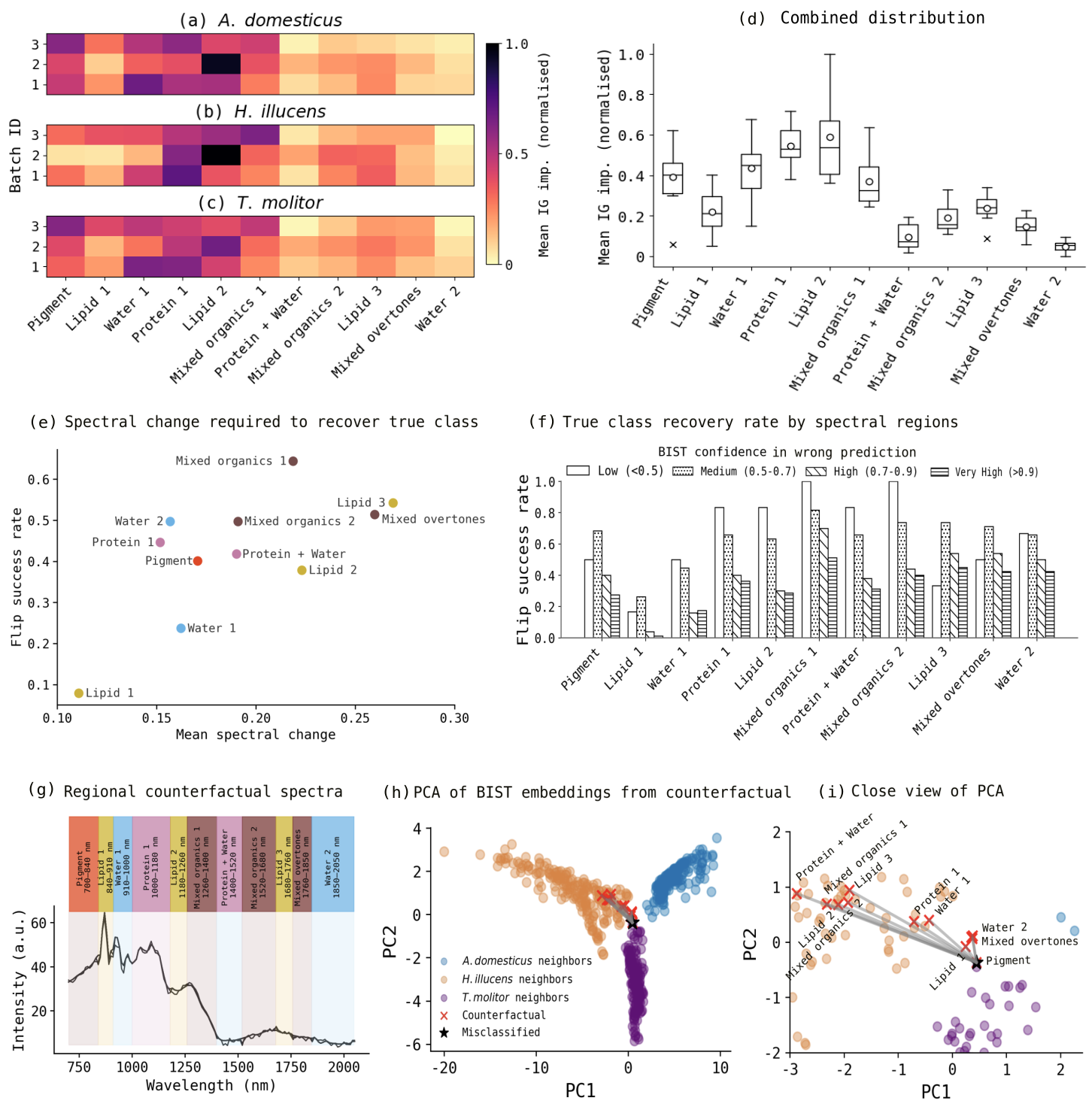}
\caption{Spectral attribution and counterfactual analysis of BISN predictions.
\textbf{(a--c)} Mean Integrated Gradients (IG) attribution per spectral region
for each species.
\textbf{(d)} IG attribution distributions across all LOBO folds.
\textbf{(e)} Mean perturbation magnitude per spectral region
against true-class recovery rate, defined as the proportion of misclassified
samples across all three LOBO folds for which optimising only that region
recovered the correct species prediction.
\textbf{(f)} True-class recovery rate per spectral region, stratified by
confidence in the wrong prediction. 
\textbf{(g)} Regional counterfactual spectra for a representative
misclassified sample (\textit{H.~illucens} predicted as \textit{T.~molitor}
with the lowest confidence in the wrong prediction). Coloured bands indicate
the perturbed region; all remaining wavelengths retain their original
spectral intensity values.
\textbf{(h)} PCA projection of BISN latent embeddings indicating the
counterfactual outcome for the same sample with the original misclassified point
(black $\star$), eleven regional counterfactuals (red $\times$), and
nearest training-set neighbours from each species.
\textbf{(i)} Zoomed view with directional arrows indicating the
embedding-space displacement induced by each regional counterfactual.
Protein~1, water~1, mixed-organics~1, and mixed-organics~2 counterfactuals
displace the point into the \textit{H.~illucens} cluster; pigment,
lipid~1, and water~2 counterfactuals fail to reach it.}
\label{fig:explain_bisn}
\end{figure*}

\vspace{0.3cm}
\noindent\textit{(a) Spectral region-level attributions}\\
We applied Integrated Gradients
(IG)~\cite{sundararajan2017axiomatic} to the $300$ held-out spectra per species in each LOBO fold (top row of Fig.~\ref{fig:explain_bisn}a-d).
Regional attribution was calculated as the mean absolute IG across wavelengths within each region, thereby accounting for unequal region widths.
Across all species and folds, IG identified a consistent set of
high-attribution regions: lipid~2 (combined mean
$0.59$), protein~1 ($0.55$),
water~1 ($0.44$), pigment
($0.39$), and mixed organics~1
($0.37$) form a clearly separated
high-attribution group with median values above $0.30$. The remaining
six regions fall around $0.20$ in median importance. These top-ranked
regions correspond directly to the dominant compositional contrasts
between species. The C--H combination bands in lipid~2 reflect the
substantially different fat fractions across species ($13$--$23\,\%$ for \textit{A.~domesticus} \cite{Laroche2019}, $12$--$43\,\%$ for \textit{T.~molitor}
\cite{MuozSeijas2025}, and
$29$--$43\,\%$ for \textit{H.~illucens}~\cite{Lawal2022, Borrelli2021}), while protein~1 captures
species-specific N--H overtone contrasts from protein fractions ranging
from $37\,\%$ to $65\,\%$ dry weight~\cite{Rumpold2013}.

The protein~1 attribution clusters tightly across folds for all
species (SD $=0.10$), confirming batch-stable reliance on this region.
In contrast, lipid~2 shows greater fold-to-fold variability (SD~$= 0.22$), driven by elevated values in \textit{A.~domesticus} and \textit{H.~illucens}. Pigment region attribution is highest for \textit{A.~domesticus} ($0.49$), followed by \textit{T.~molitor} ($0.45$) and \textit{H.~illucens} ($0.22$), reflecting species-specific differences in surface pigmentation and cuticle composition. Whether the regions are sufficient to alter model predictions under region-restricted optimisation for species discrimination is assessed through counterfactual optimisation.

\vspace{0.3cm}
\noindent\textit{(b) Counterfactual Analysis}\\
To test whether individual spectral regions are sufficient to alter model predictions beyond attribution analysis, we applied region-constrained
counterfactual optimisation to all misclassified spectra across the three LOBO
folds. For each of the eleven chemically defined regions and each misclassified
sample, the minimal $\ell_2$-regularised perturbation restricted to that region
was optimised over 1,000 gradient steps to recover the correct species prediction.

Mixed-organics~1 ($1260$--$1400$\,nm), lipid~3 ($1680$--$1760$\,nm), and
protein--water ($1400$--$1520$\,nm) achieve the highest flip success rates
($>0.40$) at relatively small perturbation magnitudes
(Fig.~\ref{fig:explain_bisn}e--f), confirming that targeted
modifications within these windows are sufficient to correct prediction errors.
Water~1 ($910$--$1000$\,nm) and lipid~1 ($840$--$910$\,nm) show flip rates
below $0.30$, indicating that even perturbations of 
comparable magnitude to other regions cannot alter the model's decision.
This discrepancy between high IG attribution and
low flip success rate for lipid~1 indicates that the region contributes to
correct predictions under normal conditions but loses its discriminative
sufficiency when PAW-induced lipid oxidation has already perturbed the C--H
profile. Mixed-organics~1, by contrast, encodes overlapping C--H and N--H
combination bands, providing redundant
biochemical signal that remains effective under single-treatment perturbations. It demonstrates that IG and counterfactual optimisation showed partial agreement. IG summarised path-based attribution across all held-out spectra, while counterfactual optimisation evaluated region-restricted changes only for misclassified spectra. Therefore, low mean IG does not imply that a region cannot alter predictions under an optimised intervention.

A representative misclassified \textit{H.~illucens} spectrum, incorrectly
assigned to \textit{T.~molitor} with
the lowest confidence (i.e., closest to random),
illustrates this structure
(Fig.~\ref{fig:explain_bisn}g--i). Perturbations restricted to
protein~1, water~1, and mixed-organics regions shift the embedding into the
correct \textit{H.~illucens} cluster, consistent with the distinct N--H and
O--H absorption signatures arising from the compositional differences between
these two species in the $1000$--$1520$\,nm interval. Perturbations in
pigment, lipid~1, and water~2 fail to reach the \textit{H.~illucens} cluster.

From the experiments, we can observe that the IG and counterfactual optimisation provided complementary rather than identical evidence. Their partial agreement reflects their different purposes: IG quantifies average path-based attribution across held-out spectra, whereas counterfactual optimisation tests whether region-restricted changes can correct individual misclassifications.

\subsection{Ablation Study}
\label{sec:ablation}

\begin{table*}[t]
\centering
\caption{Ablation study results.
\textbf{(a)} BISN component ablations.
\textbf{(b)} Effect of adding treatment and ultrasound metadata to the frozen BISN
embedding.
Convergence comparison was defined by the number of 
epochs required to reach the best validation performance (post-hoc). All values are mean $\pm$ standard deviation across three folds.}
\label{tab:ablation_bisn}
\begin{tabular}{lcc}
\toprule
\textbf{Configuration} & \textbf{Acc} & \textbf{Epochs to converge} \\
\midrule
\multicolumn{3}{l}{\textit{(a) BISN component ablations}} \\
\midrule
BISN & $0.93 \pm 0.04$ & $120 \pm 11$ \\
SG--initialised convolution weights replaced with random weights & $0.93 \pm 0.04$ & $175 \pm 15$ \\
No learnable preprocessing (raw input to encoder) & $0.73 \pm 0.12$ & $90 \pm 13$ \\
No learnable preprocessing (classical preprocessing to encoder) & $0.86 \pm 0.01$ & $99 \pm 10$ \\
No entropy-regularised batch auxiliary branch ($\lambda(e) = 0$ in Eq.~\ref{eq:joint_loss}) & $0.87 \pm 0.09$ & $125 \pm 13$ \\
\midrule
\multicolumn{3}{l}{\textit{(b) Metadata augmentation of frozen BISN embedding ($\mathbf{z}$)}} \\
\midrule
$\text{BISN embedding}$ only & $0.93 \pm 0.04$ & $34 \pm 5$ \\
$[\text{BISN embedding};\,\text{treatment}]$ & $0.93 \pm 0.04$ & $41 \pm 3$ \\
$[\text{BISN embedding};\,\text{ultrasound}]$ & $0.93 \pm 0.04$ & $38 \pm 3$ \\
$[\text{BISN embedding};\,\text{treatment};\,\text{ultrasound}]$ & $0.93 \pm 0.04$ & $40 \pm 5$ \\
\bottomrule
\end{tabular}
\end{table*}

Two sets of ablation experiments were designed to isolate the contribution of each BISN component and to test
whether explicit treatment and ultrasound metadata improve species classification from the frozen embedding (Table~\ref{tab:ablation_bisn}).
Removing the learnable preprocessing module reduces mean LOBO accuracy from
$0.93 \pm 0.04$ to $0.73 \pm 0.12$, the largest single drop in the ablation
table, establishing adaptive spectral preprocessing as the primary mechanism
enabling cross-batch generalisation. Replacing it with fixed optimal classical
preprocessing recovers accuracy to $0.86 \pm 0.01$, seven percentage points
below standard BISN. It confirms that a batch-agnostic filter cannot match a module
jointly trained to suppress batch-correlated structure while preserving
species-discriminative variation.
Replacing SG-informed kernel weights with random initialisation does not alter accuracy ($0.93 \pm 0.04$) but increases convergence time by
approximately $55$ epochs ($175 \pm 15$ vs. $120 \pm 11$). The SG initialisation places the convolution near an effective solution and accelerates convergence without constraining the optimised kernels to retain the analytic SG shape (Supplementary~\ref{app:learned_kernels}, Fig.~\ref{fig:learned_sg_kernels}).

Removing the batch auxiliary branch ($\lambda(e) = 0$) reduces accuracy to $0.87 \pm 0.09$,
with the fold-to-fold standard deviation more than doubled ($0.09$ vs. $0.04$).
This configuration achieves accuracy similar 
to TabNet ($0.88 \pm 0.06$), indicating that without batch entropy regularisation, the 
preprocessing alone does not substantially improve over strong baselines. 
Without it, the preprocessing module may exploit
batch-correlated structure to minimise species cross-entropy, producing a
representation that generalises poorly across batches. This is consistent with
the elevated batch-probe accuracy observed in the BISN preprocessing output
without batch auxiliary regularisation.

Furthermore, 
adding ground-truth treatment and ultrasound labels to the frozen BISN embedding did not improve species accuracy
($0.93 \pm 0.04$ in all cases). 
Thus, explicit metadata provided no additional predictive value for species classification.
This is consistent with the MANOVA
results in Section~\ref{sec:representation_analysis}, where treatment and
ultrasound effects remain statistically detectable in the embedding alongside
the dominant species signal.
The results demonstrate that the SG initialisation improves training efficiency
without affecting final accuracy. 
The learnable preprocessing and the auxiliary entropy branch contributed to BISN performance.
While SG initialisation accelerates convergence, removing any single component
degrades both accuracy and cross-batch stability toward baseline levels.

\section{Conclusions}
\label{sec:conclusion}

We presented Batch-Invariant Spectral Network (BISN) for robust cross-batch insect species
authentication from near-infrared spectra. BISN applies batch auxiliary entropy regularisation to the preprocessing output before the sparse attentive encoder. 
We evaluated under a strict
leave-one-batch-out protocol on $2{,}700$ spectra from three edible
insect species across three production batches, treatments, and ultrasound
conditions. BISN achieved $0.93 \pm 0.04$ accuracy in classifying insect species from milled-slurry spectra, exceeding the best baseline by four percentage points. Our experiments confirmed that the learnable preprocessing and batch auxiliary branch jointly contributed to this gain.
Integrated Gradients highlighted lipid and protein regions, while counterfactual optimisation identified mixed-organics, lipid~3, and protein--water regions as effective intervention targets for correcting misclassifications. Their partial agreement provides complementary evidence about BISN attribution and decision-boundary sensitivity.

Despite these results, this study has several limitations.
Each batch represented a separate commercial purchase and was measured on a different day. This design reflects supply-chain shifts where lot composition, storage history, preparation, acquisition day, and instrument state may change together. The evaluation therefore tests their combined effect but cannot separate biological variation from temporal instrumental drift. Future studies should cross biological batches with repeated acquisition days, reference standards, and multiple instruments.

Moreover, the pigment region 
received consistently high IG attribution but 
showed the largest inter-fold variability,
suggesting scattering-dominated contrast that
may be less stable than protein and lipid absorption bands under
instrument or sampling variation. The reliability of this region as an
authentication cue in operational settings requires further validation.
Future work should expand the species panel and batch count, and evaluate BISN under
semi-supervised or few-shot adaptation scenarios where limited unlabelled
target spectra are available at deployment time.
The internal validation split supported checkpoint selection but did not reproduce a new-batch shift within the two training batches. Future studies with more independent batches should use nested LOBO validation. Validation under inline industrial acquisition remains future work.

\vspace{0.3cm}
\noindent\textbf{CRediT authorship contribution statement}\\
\textbf{Majharulislam Babor:} Conceptualization, Formal analysis, Methodology, Software, Visualization, Writing – original draft;
\textbf{Giacomo Rossi:} Data curation, Investigation, Methodology, Resources, Writing – review \& editing;
\textbf{Annalisa Altavilla:} Data curation, Investigation;
\textbf{Oliver Schlüter:} Funding acquisition, Project administration;
\textbf{Marina M.-C. Höhne:} Conceptualization, Methodology, Supervision, Visualization, Writing – review \& editing.

\vspace{0.3cm}
\noindent\textbf{Data availability}\\
The data supporting the investigations and findings of this study are publicly available at\\ \href{https://github.com/majharB/bisn/tree/main/data} {https://github.com/majharB/bisn/tree/main/data}

\vspace{0.3cm}
\noindent\textbf{Declaration of competing interest}\\
The authors declare that they have no known competing financial
interests or personal relationships that could have appeared to influence
the work reported in this paper.

\vspace{0.3cm}
\noindent\textbf{Funding}\\
This work was supported by the Federal Ministry of Food and Agriculture (BMEL), Germany through the Federal Office for Agriculture and Food (BLE) under
the innovation support programme, grant no.\ 281A809A21, within the
research project ProtinA (\textit{Ern\"{a}hrungs-optimierte
Erschlie\ss{}ung alternativer Proteinquellen durch innovative und
nachhaltige Verarbeitungstechnologien am Beispiel von Grillen Acheta
domestica}).


\end{multicols}

\appendix
\setcounter{table}{0}
\setcounter{figure}{0}

\renewcommand{\thetable}{S.\arabic{table}}
\renewcommand{\thefigure}{S.\arabic{figure}}
\section*{Supplementary Materials}
\renewcommand{\thesection}{S\arabic{section}}
\setcounter{section}{0}

\section{Experimental Dataset}
\label{sec:dataset}

The allocation of samples across all batch, treatment, and ultrasound combinations is shown in Table~\ref{tab:factorial_design}.
Here, “Samples = 50” denotes 50 distinct physical subsamples for each batch $\times$ species $\times$ treatment $\times$ ultrasound combination. Each subsample was taken from the corresponding processed batch condition and measured once by NIR spectroscopy. The replication therefore reflects variation among physical subsamples within a factorial cell, rather than repeated scans of the same prepared specimen. Here, an example of a factorial cell is batch 1 of \textit{A. domesticus} with fresh water (T0) and no ultrasound (U0).

\begin{table*}[htbp]
\centering
\caption{Factorial design of the dataset showing the number of spectra per
insect species, production batch, processing treatment, and ultrasound condition.
Each factorial cell contains $n = 50$ replicate spectra.
T0: fresh insects in tap water; T1: blanched at $70\,^\circ\mathrm{C}$, 5\,min;
T2: plasma-activated water.
U0: no ultrasound; U1: ultrasound applied.}
\label{tab:factorial_design}
\small
\begin{tabular}{llllc}
\toprule
\textbf{Production Batch} & \textbf{Insect Species} & \textbf{Treatment} & \textbf{Ultrasound} & \textbf{Samples} \\
\midrule
\multirow{18}{*}{B1 ($n=900$)}
& \multirow{6}{*}{\textit{A.~domesticus} ($n=300$)}
& T0 & U0 & 50 \\
& & T1 & U0 & 50 \\
& & T2 & U0 & 50 \\
& & T0 & U1 & 50 \\
& & T1 & U1 & 50 \\
& & T2 & U1 & 50 \\
\cmidrule(lr){2-5}
& \multirow{6}{*}{\textit{H.~illucens} ($n=300$)}
& T0 & U0 & 50 \\
& & T1 & U0 & 50 \\
& & T2 & U0 & 50 \\
& & T0 & U1 & 50 \\
& & T1 & U1 & 50 \\
& & T2 & U1 & 50 \\
\cmidrule(lr){2-5}
& \multirow{6}{*}{\textit{T.~molitor} ($n=300$)}
& T0 & U0 & 50 \\
& & T1 & U0 & 50 \\
& & T2 & U0 & 50 \\
& & T0 & U1 & 50 \\
& & T1 & U1 & 50 \\
& & T2 & U1 & 50 \\
\midrule
B2 ($n=900$) & \multicolumn{4}{c}{identical structure as B1} \\
\midrule
B3 ($n=900$) & \multicolumn{4}{c}{identical structure as B1} \\
\bottomrule
\end{tabular}
\end{table*}

\section{Sparse Attentive Encoder}
\label{app:encoder_structure}

\begin{figure*}[t]
\centering
\includegraphics[width=\textwidth]{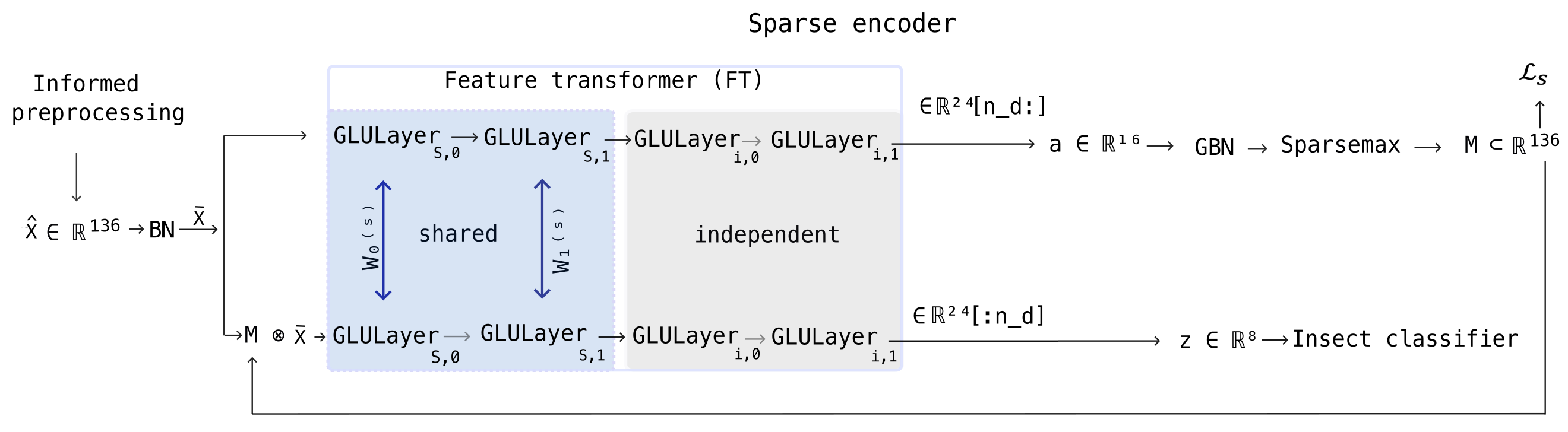}
\caption{Internal structure of the sparse attentive encoder (Component~2).
The batch-normalised input $\bar{\mathbf{x}} \in \mathbb{R}^{136}$
follows two parallel paths.
Attention path (top): $\bar{\mathbf{x}}$ passes through
$\mathrm{FT}_{\mathrm{split}}$ ($n_s=2$ shared + $n_i=2$ independent
GLU layers); the last $n_a=16$ dimensions form the attention seed,
projected to $\mathbb{R}^d$ via $\mathbf{W}_a$, normalised by GBN,
and passed through sparsemax to yield the sparse mask
$\mathbf{M} \in \Delta^{d-1}$.
Feature extraction path (bottom): $\mathbf{M} \otimes \bar{\mathbf{x}}$
produces the masked spectrum $\bar{\mathbf{x}}_{\mathrm{sel}}$.
Embedding path (bottom): $\bar{\mathbf{x}}_{\mathrm{sel}}$
passes through $\mathrm{FT}_{\mathrm{step}}$, which shares weight
matrices $\mathbf{W}_0^{(s)}$, $\mathbf{W}_1^{(s)}$ with
$\mathrm{FT}_{\mathrm{split}}$ (double-headed arrows); the first
$n_d=8$ output dimensions yield the
embedding $\mathbf{z} \in \mathbb{R}^{8}$.
All residual additions are scaled by $1/\sqrt{2}$~\citep{arik2021tabnet};
GBN: Ghost Batch Normalisation~\citep{hoffer2017}.}
\label{fig:encoder_structure}
\end{figure*}

The internal structure of
the sparse attentive encoder (Component~2, Section~\ref{sec:bisn}) is presented in Fig.~\ref{fig:encoder_structure}. The $\mathrm{FT}_{\mathrm{split}}$ and $\mathrm{FT}_{\mathrm{step}}$
share the first two GLU layers while maintaining independent layers
for wavelength-mask generation and latent feature extraction,
respectively.

\noindent\textbf{Feature transformer.}
Both $\mathrm{FT}_{\mathrm{split}}$ and $\mathrm{FT}_{\mathrm{step}}$
map their input to $\mathbb{R}^{f}$ ($f = n_d + n_a = 24$)
through four sequential residual GLU layers.
The first two layers carry shared weight matrices
$\mathbf{W}^{(s)}_0 \in \mathbb{R}^{2f \times d}$ and
$\mathbf{W}^{(s)}_1 \in \mathbb{R}^{2f \times f}$ across both
transformers; the remaining two are transformer-specific.
For input $\mathbf{u} \in \mathbb{R}^d$:
\[
\begin{aligned}
\mathbf{h}_0 &= \mathrm{GLULayer}_{s,0}(\mathbf{u}), \\
\mathbf{h}_1 &= \tfrac{1}{\sqrt{2}}\bigl(\mathbf{h}_0
 + \mathrm{GLULayer}_{s,1}(\mathbf{h}_0)\bigr), \\
\mathbf{h}_2 &= \tfrac{1}{\sqrt{2}}\bigl(\mathbf{h}_1
 + \mathrm{GLULayer}_{i,0}(\mathbf{h}_1)\bigr), \\
\mathrm{FT}(\mathbf{u}) &= \tfrac{1}{\sqrt{2}}\bigl(\mathbf{h}_2
 + \mathrm{GLULayer}_{i,1}(\mathbf{h}_2)\bigr),
\end{aligned}
\]
where $s$ and $i$ denote shared and independent blocks.
The first layer receives $d=136$; all subsequent layers receive
$\mathbb{R}^{f}$ from the preceding layer.

\noindent\textbf{Sparse wavelength mask.}
$\mathrm{FT}_{\mathrm{split}}$ processes $\bar{\mathbf{x}}$ and
produces a joint representation in $\mathbb{R}^{24}$.
The last $n_a = 16$ dimensions form the attention seed $\mathbf{a}$,
which is projected and normalised:
\[
\tilde{\mathbf{a}} = \mathrm{GBN}(\mathbf{W}_a\,\mathbf{a}),
\qquad
\mathbf{a} = \mathrm{FT}_{\mathrm{split}}(\bar{\mathbf{x}})_{[n_d:]},
\qquad
\mathbf{W}_a \in \mathbb{R}^{d \times n_a}.
\]
The sparse mask is obtained by applying sparsemax
(Supplementary~\ref{app:sparsemax}):
\[
\mathbf{M}
= \mathrm{sparsemax}(\tilde{\mathbf{a}}) \in \Delta^{d-1},
\qquad M_i \geq 0,
\quad \textstyle\sum_{i=1}^{d} M_i = 1.
\]
The masked spectrum
$\bar{\mathbf{x}}_{\mathrm{sel}} = \mathbf{M} \otimes \bar{\mathbf{x}}$
fully suppresses wavelengths assigned zero weight by sparsemax.

\noindent\textbf{Latent embedding.}
$\mathrm{FT}_{\mathrm{step}}$ processes $\bar{\mathbf{x}}_{\mathrm{sel}}$;
the first $n_d$ output dimensions yield the signed 
latent embedding:

\[
\mathbf{z}
=
\mathrm{FT}_{\mathrm{step}}
\bigl(\bar{\mathbf{x}}_{\mathrm{sel}}\bigr)_{[:n_d]}
\in \mathbb{R}^{8}.
\]

This embedding forms the basis for the Integrated Gradients and
counterfactual analyses (Section~\ref{sec:explain_bisn}).

\subsection{Sparsemax}
\label{app:sparsemax}

Sparsemax~\cite{martins2016} projects a raw score vector
$\mathbf{u} \in \mathbb{R}^d$ onto the probability simplex
$\Delta^{d-1}$ (Supplementary~\ref{app:probability_simplex}) by solving
\begin{equation}
 \mathrm{sparsemax}(\mathbf{u})
 = \argmin_{\mathbf{s}\,\in\,\Delta^{d-1}} \|\mathbf{s} - \mathbf{u}\|^2.
 \label{eq:sparsemax}
\end{equation}
The Karush-Kuhn-Tucker (KKT) conditions of this constrained quadratic programme give
$s_i^\star = \max(u_i - \lambda^\star,\, 0)$, where $\lambda^\star$
enforces $\sum_i s_i^\star = 1$.
Sorting scores in descending order $u_{(1)} \geq \cdots \geq u_{(d)}$,
the support size $\rho$ is the largest integer satisfying
\begin{equation}
 1 + \rho\, u_{(\rho)} > \sum_{j=1}^{\rho} u_{(j)},
 \label{eq:support_cond}
\end{equation}
and the threshold is
\begin{equation}
 \tau(\mathbf{u})
 = \frac{\displaystyle\sum_{j=1}^{\rho} u_{(j)} - 1}{\rho}.
 \label{eq:tau}
\end{equation}
The closed-form solution is therefore
\begin{equation}
 \mathrm{sparsemax}(\mathbf{u})_i
 = \max\!\bigl(u_i - \tau(\mathbf{u}),\; 0\bigr).
 \label{eq:sparsemax_solution}
\end{equation}
Wavelengths whose score falls below $\tau$ receive weight exactly zero
and are excluded from downstream computation.
Because sparsemax is differentiable, its Jacobian for backpropagation
over the active support $\mathcal{S} = \{i: \mathrm{sparsemax}(\mathbf{u})_i > 0\}$ is
\begin{equation}
 \frac{\partial\,\mathrm{sparsemax}(\mathbf{u})_i}{\partial u_j}
 = \begin{cases}
 \delta_{ij} - \dfrac{1}{|\mathcal{S}|}
 & i \in \mathcal{S},\; j \in \mathcal{S}, \\[4pt]
 0 & \text{otherwise,}
 \end{cases}
 \label{eq:sparsemax_grad}
\end{equation}
where $\delta_{ij}$ is the Kronecker delta.

\subsection{Probability Simplex}
\label{app:probability_simplex}

The $(d{-}1)$-dimensional probability simplex is
\begin{equation}
 \Delta^{d-1}
 = \bigl\{\,\mathbf{s} \in \mathbb{R}^d
 \;\big|\;
 s_i \geq 0 \;\; \forall\, i,\;\;
 \textstyle\sum_{i=1}^d s_i = 1
 \,\bigr\}.
 \label{eq:simplex}
\end{equation}
Softmax always maps to the interior of $\Delta^{d-1}$, assigning
strictly positive weight to every element.
Sparsemax can map to the boundary of the space, where one or more weights are
exactly zero.

\subsection{Integrated Gradients}
\label{app:ig_derivation}

With the zero-spectrum baseline ($\mathbf{x}'=\mathbf{0}$), the
Integrated Gradients attribution of input dimension $i$ for class $c$ is
{

\begin{equation}
 \mathrm{IG}_i^c(\mathbf{x})
 =
 (x_i-x_i')
 \int_0^1
 \frac{
 \partial \hat{y}_c
 \left(
 \mathbf{x}'
 +\alpha(\mathbf{x}-\mathbf{x}')
 \right)
 }{
 \partial x_i
 }
 \,\mathrm{d}\alpha,
 \qquad
 \mathbf{x}'=\mathbf{0}.
\label{eq:ig_general}
\end{equation}
}
The integral is approximated with:
{
\begin{equation}
\mathrm{IG}_i^c(\mathbf{x})
\approx
(x_i-x_i')
\sum_{t=1}^{T}
w_t
\frac{
\partial \hat{y}_c
\left(
\mathbf{x}'
+\alpha_t(\mathbf{x}-\mathbf{x}')
\right)
}{
\partial x_i
},
\qquad
\mathbf{x}'=\mathbf{0},
\quad
T=1000.
\label{eq:ig_approx}
\end{equation}
}

\noindent 
Here, $\alpha_t$ and $w_t$ denote the Gauss--Legendre nodes and weights mapped to the interval $[0,1]$.
The numerical accuracy of the Integrated Gradients approximation was verified
for each sample using the completeness property:
$\sum_i \mathrm{IG}_i^c \approx \hat{y}_c(\mathbf{x}) - \hat{y}_c(\mathbf{0})$
with an absolute reconstruction error below $10^{-3}$.

\clearpage
\section{ Model architectures and training details}
\label{sec:training_details}
\paragraph{BISN.}
All BISN components were optimised jointly using Adam for up to 200 epochs.
The contribution of the batch auxiliary branch was annealed from zero to one following
Ganin et al.~\cite{ganin2016dann}:
\begin{equation}
 \lambda(e) = \frac{2}{1 + \exp(-10 \cdot e / E)} - 1,
 \label{eq:annealing}
\end{equation}
where $e$ is the current epoch and $E = 200$ is the total number of epochs.
This schedule prevents the Batch auxiliary branch from destabilising
species classification in early epochs before stable latent representations
have emerged. The parameter-wise gradient routing
is summarised in Table~\ref{tab:gradient_routing}.

{
\paragraph{DANN architecture.}
The domain-adversarial neural network (DANN) used a multilayer perceptron feature extractor. The 136-dimensional input passed through layers containing 128, 128, and 64 units. Each layer used batch normalisation and ReLU activation, with dropout (\(p=0.20\)) after the first two layers. A linear \(64\!-\!3\) classification head predicted the three insect species. The domain branch applied a gradient reversal layer to the 64-dimensional embedding. It comprised a \(64\!-\!32\!-\!2\) multilayer perceptron with ReLU activation and dropout (\(p=0.20\)) after the hidden layer. The two outputs represented the two training batches within each LOBO fold.
}
Hyperparameter search spaces for BISN and the baseline models are provided in
Table~\ref{tab:all_hyperparameters}.

\begin{table*}[htbp]
\centering
\caption{Gradient routing in Batch-Invariant Spectral Network (BISN).
Each row shows a parameter group, the effective loss signal it receives, and the
resulting update direction.
The sign of the \(\mathcal{L}_{\mathrm{b}}\) term reaching \(\theta_x\) is reversed
by the gradient reversal layer, driving the preprocessing module to suppress
batch-identifying structure.}
\label{tab:gradient_routing}
\small
\begin{tabular}{llll}
\toprule
Parameters & Module & Effective loss signal & Direction \\
\midrule
$\theta_b$ & Batch auxiliary head (component 4) & $\lambda(e)\,\mathcal{L}_b$ & minimise ($= $ increase auxiliary entropy) \\

$\theta_y$ & Species classifier (component 3) & $\mathcal{L}_y$ & minimise species cross-entropy\\

$\theta_f$ & Sparse encoder (component 2) & $\mathcal{L}_y + \beta\,\mathcal{L}_s$ & minimise \\

$\theta_x$ & Informed preprocessing (component 1) & $\mathcal{L}_y + \beta\,\mathcal{L}_s - \lambda(e)\,\mathcal{L}_b$ & minimise (via reversed gradient) \\
\bottomrule
\end{tabular}
\end{table*}

\begin{table*}[ht]
\centering
\caption{The BISN preprocessing parameters were obtained from the classical preprocessing grid search. All other parameters marked "Tuned" were selected by five-fold stratified cross-validation maximising species accuracy on the training batches, where bold entries within search space indicate the selected optimal value per parameter. 
Values marked "Fixed" are determined by the experimental design or the reference implementation and were not subject to search.}
\label{tab:all_hyperparameters}
\small
\begin{tabular}{@{}llp{5.5cm}l@{}}
\toprule
\textbf{Model/ Module} & \textbf{Parameter} & \textbf{Search space} & \textbf{Selection} \\
\midrule
\textit{(a) BISN} & &\\
\midrule
\multirow{3}{*}{Preprocessing}
 & SG window size $w$ & $\{11, 21, 31, 41, 51, \mathbf{61}, 71\}$ & Tuned \\
 & Polynomial degree $r$ & $\{\mathbf{1}, 2, 3\}$ & Tuned \\
 & Derivative order $k$ & $\{0, \mathbf{1}, 2\}$ & Tuned \\
\midrule

\multirow{6}{*}{Sparse encoder}
 & Shared FT blocks $n_s$ & $\{\mathbf{2}, 3\}$ & Tuned \\
 & Independent FT blocks $n_i$ & $\{\mathbf{2}, 3\}$ & Tuned \\
 & Embedding dimension $n_d$ & $\{4, \mathbf{8}, 16, 32\}$ & Tuned \\
 & Attention dimension $n_a$ & $\{8, \mathbf{16}, 32, 64\}$ & Tuned \\
 & Output dimension & $\{\mathbf{8}, 16\}$ & Tuned \\
 & Decision steps $n_{\mathrm{steps}}$ & 1 & Fixed \\
 & Numerical stability constant $\epsilon$ & $10^{-12}$ & Fixed \\
\midrule

\multirow{4}{*}{Batch Auxiliary Branch}
 & Virtual mini-batch size $v$ & $\{32, \mathbf{64}\}$ & Tuned \\
 & Auxiliary weight $\lambda(e)$ & $[0,1]$ (scheduled by Eq. \ref{eq:annealing}) & Fixed \\
 & Sparse-mask entropy weight $\beta$ & $\{10^{-4}, \mathbf{10^{-3}}, 10^{-2}, 10^{-1}\}$ & Tuned \\
\midrule

\multirow{4}{*}{Optimisation}
 & Optimiser & Adam & Fixed \\
 & Learning rate $\eta$ & $\{10^{-4}, \mathbf{10^{-3}}, 5\times10^{-3}\}$ & Tuned \\
 & Mini-batch size & $\{4, \mathbf{8}, 16, 32, 64\}$ & Tuned \\
 & Training epochs $E$ & $\{100, \mathbf{200}, 250\}$ & Tuned \\
\midrule
\textit{(b) Baseline Models}\\
\midrule

\multirow{2}{*}{LDA~\cite{fisher1936use}}
 & Solver & \textbf{LSQR}, SVD & Tuned \\
 & Shrinkage & Automatic & Fixed \\
\midrule

\multirow{2}{*}{GPC~\cite{rasmussen2006gaussian}}
 & Kernel length-scale & $\{0.1,\, 0.5,\, \mathbf{1.0},\, 2.0,\, 5.0,\, 10.0\}$ & Tuned \\
 & Optimiser restarts & $5$ & Fixed \\
\midrule

\multirow{2}{*}{PLSDA~\cite{barker2003partial}}
 & Latent variables & $\{2, 3, 5, 10, 15, \mathbf{20}, 25\}$ & Tuned \\
 & Algorithm & NIPALS & Fixed \\
\midrule

diPLS~\cite{nikzad2018domain}
 & Latent variables & Automatic for 3 folds $\{ 24,\, 20,\, 18\}$ & Tuned\\
\midrule

\multirow{3}{*}{PDS-PLSDA~\cite{wang1991multivariate}}
 & Window width & $\{ 11,\, 13,\, 15,\, \mathbf{21},\, 23\}$ & Tuned \\
 & Overlap & $\{0.3,\, \mathbf{0.5},\, 0.7\}$ & Tuned \\
 & Latent variables & $\{10,\, 15,\, \mathbf{20},\, 25\}$ & Tuned \\
\midrule

ShapDA~\cite{Babor2026ShapDA}
 & Latent dimension & $\{8,\, 10,\, 12,\, 14,\, 16,\, 18,\, 20,\, 25,\, 30\}$ & Tuned \\
\midrule

\multirow{3}{*}{SpectraTr~\cite{fu2022spectratr}}
 & Learning rate & $\{\mathbf{10^{-3}},\, 10^{-4},\, 10^{-5}\}$ & Tuned \\
 & Weight decay & $\{0,\, \mathbf{10^{-4}},\, 10^{-3}\}$ & Tuned \\
 & Batch size & $\{4,\, \mathbf{8},\, 16,\, 32,\, 64\}$ & Tuned \\
\midrule

\multirow{3}{*}{NIRCoreVision-MLP~\cite{singh2025nircorevision}}
 & Latent dimension & $\{8,\, 12,\, 16,\, \mathbf{20},\, 30\}$ & Tuned \\
 & Core-set fraction & $\{0.05, 0.1, \mathbf{0.15}, 0.30\}$ & Tuned \\
 & Learning rate & $\{\mathbf{10^{-3}},\, 10^{-4},\, 10^{-5}\}$ & Tuned \\
\midrule

TabPFN~\cite{hollmann2025accurate}
 & Prior fits & 25 & Fixed \\
\midrule

\multirow{4}{*}{TabNet~\cite{arik2021tabnet}}
 & Decision steps & $\{\mathbf{1},\, 2,\, 3\}$ & Tuned \\
 & Attention dimension & $\{8,\, \mathbf{16},\, 32,\, 64\}$ & Tuned \\
 & Decision dimension & $\{\mathbf{8},\, 16,\, 32,\, 64\}$ & Tuned \\
 & Sparsity coefficient & $\{10^{-8},\, \mathbf{10^{-5}},\, 10^{-4},\, 10^{-3}\}$ & Tuned \\
\midrule

\multirow{5}{*}{DANN~\cite{ganin2016dann}}
 & Learning rate & $\{\mathbf{10^{-3}},\, 10^{-4},\, 10^{-5}\}$ & Tuned \\
 & Dropout rate & $\{0.1,\, \mathbf{0.2},\, 0.3\}$ & Tuned \\
 & Batch size & $\{4,\, \mathbf{8},\, 16,\, 32,\, 64\}$ & Tuned \\
 & Epochs & $200$ & Fixed \\
 & Weight decay & $10^{-4}$ & Fixed \\
\bottomrule
\end{tabular}
\end{table*}

\clearpage
\section{Interpretation of Spectral Wavelength Regions}
\label{app:wavelength_regions}

To facilitate interpretation of the learned spectral representations, the NIR range
used in this study (700--2050\,nm) was partitioned into eleven chemically
interpretable intervals based on known overtone and combination bands of major
biochemical constituents in biological tissues and food
matrices~\cite{burns2007handbook,tsenkova2009aquaphotomics,swir_review_2015,food_nir_review_2025}.
Table~\ref{tab:wavelength_segments} lists each interval, its chemical interpretation,
and the spectroscopic justification.

\begin{table*}[htbp]
\centering
\caption{Spectral regions of the NIR range used in this study and their
biochemical interpretations based on established spectroscopy literature.}
\label{tab:wavelength_segments}
\small
\begin{tabular}{p{0.13\textwidth} p{0.16\textwidth} p{0.71\textwidth}}
\toprule
\textbf{Range (nm)} & \textbf{Interpretation} & \textbf{Spectroscopic justification} \\
\midrule

700--840 &
Pigment &
Transition between visible and NIR wavelengths dominated by weak overtone bands and
scattering effects.
Absorption arises primarily from pigments and chromophores rather than vibrational
transitions of the major food-matrix constituents~\cite{burns2007handbook,swir_review_2015}. \\[4pt]

840--910 &
Lipid 1 &
Features near 840--855\,nm and around 910\,nm correspond to third overtones of C--H
stretching vibrations in lipids and
carbohydrates~\cite{burns2007handbook,food_nir_review_2025}. \\[4pt]

910--1000 &
Water 1 &
The band centred near 970\,nm represents the second overtone of the O--H stretching
vibration of water and is widely used as a marker of water content in biological
tissues~\cite{burns2007handbook,tsenkova2009aquaphotomics,swir_review_2015}. \\[4pt]

1000--1180 &
Protein 1 &
Contains N--H and C--H first-overtone vibrations associated with proteins; frequently
used in NIR models for protein and biomass
quantification~\cite{burns2007handbook,food_nir_review_2025}. \\[4pt]

1180--1260 &
Lipid 2 &
Absorption near 1190--1210\,nm corresponds to second overtones and combination bands
of C--H stretching in lipid acyl
chains~\cite{burns2007handbook,swir_review_2015}. \\[4pt]

1260--1400 &
Mixed organics 1 &
Overlapping C--H and N--H combination bands from carbohydrates, proteins, and other
organic matrix components; widely used in empirical NIR calibration
models~\cite{burns2007handbook,food_nir_review_2025}. \\[4pt]

1400--1520 &
Protein--water &
A strong O--H combination band near 1450\,nm, with neighbouring N--H and C--H
contributions from proteins and
collagen~\cite{tsenkova2009aquaphotomics,swir_review_2015}. \\[4pt]

1520--1680 &
Mixed organics 2 &
Overlapping C--H combination bands from lipids and other organic constituents;
commonly interpreted as a mixed biochemical region in biological tissue
spectroscopy~\cite{burns2007handbook,swir_review_2015}. \\[4pt]

1680--1760 &
Lipid 3 &
Strong lipid features near 1700--1760\,nm due to the first overtone of C--H
stretching in fatty acids and
triacylglycerols~\cite{burns2007handbook,swir_review_2015}. \\[4pt]

1760--1850 &
Mixed overtone &
Weaker C--H and C=O combination bands from multiple biochemical constituents; no
single dominant chromophore~\cite{burns2007handbook}. \\[4pt]

1850--2050 &
Water 2 &
A strong O--H combination band centred near 1900\,nm dominates absorption in aqueous
biological samples; beyond 2000\,nm, continued O--H combination bands maintain
water dominance with weaker contributions from other functional
groups~\cite{tsenkova2009aquaphotomics,swir_review_2015}. \\
\bottomrule
\end{tabular}
\end{table*}

\section{Best Configurations for Classical Preprocessing}
\label{sec:best_classical_preprocessing}

Optimal preprocessing configurations for each LOBO fold are summarised in
Table~\ref{tab:lobo_preprocessing_results}.
Configurations for folds 1 and 2 are identical (MSC, second-order detrend, SG
first derivative $w=61$, $r=1$), suggesting that the dominant batch artefact in
these folds has a consistent spectral character (multiplicative scatter plus
quadratic baseline curvature) that MSC and a quadratic detrend address well.
Fold~3 requires only first-order detrend without MSC, indicating a different
predominant artefact structure for batch~3, possibly related to sample preparation
differences on that acquisition day.

\begin{table*}[ht]
\centering
\caption{Optimal classical preprocessing configurations per LOBO fold.
SNV: standard normal variate; MSC: multiplicative scatter correction;
SG: Savitzky--Golay smoothing/derivative. Configurations were selected using the rank-sum criterion
described in Section~\ref{sec:preprocessing} and were applied consistently
to all baseline models evaluated in that fold.}
\label{tab:lobo_preprocessing_results}
\begin{tabular}{ccccccc}
\toprule
Left-out batch & SNV & MSC & Detrend degree & SG window & SG poly order & SG deriv order \\
\midrule
1 & False & True & 2 & 61 & 1 & 1 \\
2 & False & True & 2 & 61 & 1 & 1 \\
3 & False & False & 1 & 61 & 1 & 1 \\
\bottomrule
\end{tabular}
\end{table*}

\section{Bootstrap Comparison of BISN and DANN}
\label{app:bootstrap_comparison}

To verify the statistical significance of BISN performance over the best-performing baseline, DANN, we performed bootstrap resampling grouped by insect species to maintain species balance in the test set. Table~\ref{tab:bootstrap_comparison} reports bootstrap resampling results
(1,000,000 iterations) comparing BISN and the best-performing baseline (DANN with
classical preprocessing) on the concatenated test predictions from three LOBO
folds combined.
Confidence intervals were constructed using the percentile method.
BISN achieves statistically significant superiority over DANN on both accuracy and
F1 score, with the difference exceeding four percentage points and a two-sided
\(p\)-value below \(10^{-6}\) for both metrics. To account for the hierarchical structure of the test data, we additionally performed a paired preparation-block bootstrap within each held-out LOBO batch, using species $\times$ treatment $\times$ ultrasound preparation block as the resampling unit. BISN showed higher accuracy than DANN in all three held-out batches and the equal-weight mean accuracy difference across the three LOBO folds was statistically significant under the conditional preparation-block bootstrap ($p=0.002$). This inference is conditional on the three observed production batches and should not be interpreted as population-level significance across independent batches.

\begin{table*}[htbp]
\centering
\caption{Bootstrap comparison of BISN and DANN.
\textbf{(a)} Comparison on the concatenated test sets of all LOBO folds.
\textbf{(b)} Paired preparation-block bootstrap comparison
within each held-out LOBO batch and for the equal-weight mean across folds.
All bootstrap analyses used 1,000,000 iterations and percentile 95\%
confidence intervals. For panel (b), the resampling unit was one species
$\times$ treatment $\times$ ultrasound preparation block within each
held-out batch (18 preparation blocks per batch; 50 spectra per block).
The reported \(p\)-values are two-sided
bootstrap probabilities. Here, EWM is the equal-weight mean that assigns equal weight to
the three LOBO folds, and its confidence interval and \(p\)-value are
conditional on the three observed production batches.}
\label{tab:bootstrap_comparison}

\textbf{(a) Concatenated test sets of all LOBO folds}

\vspace{2mm}

\begin{tabular}{lccccc}
\toprule
\textbf{Metric} & \textbf{Model} & \textbf{Mean} &
\textbf{95\% CI} & \textbf{Difference (BISN $-$ DANN)} &
\textbf{\(p\)-value} \\
\midrule

\multirow{2}{*}{Acc}
 & BISN & 0.934 & [0.925, 0.944]
 & \multirow{2}{*}{0.041\ [0.023, 0.067]}
 & \multirow{2}{*}{$<10^{-6}$} \\
 & DANN & 0.893 & [0.877, 0.902] & & \\

\midrule

\multirow{2}{*}{F1 score}
 & BISN & 0.934 & [0.925, 0.944]
 & \multirow{2}{*}{0.041\ [0.024, 0.068]}
 & \multirow{2}{*}{$<10^{-6}$} \\
 & DANN & 0.893 & [0.876, 0.901] & & \\

\bottomrule
\end{tabular}

\vspace{5mm}


{

\textbf{(b) Paired preparation-block bootstrap within held-out LOBO batches}

\vspace{2mm}

\begin{tabular}{lccccc}
\toprule
\textbf{Left-out batch} & \textbf{Model} & \textbf{Acc} &
\textbf{95\% CI} & \textbf{Difference (BISN $-$ DANN)} &
\textbf{\(p\)-value} \\
\midrule

\multirow{2}{*}{1}
 & BISN & 0.934 & [0.908, 0.959]
 & \multirow{2}{*}{0.044\ [$-$0.011, 0.119]}
 & \multirow{2}{*}{0.164} \\
 & DANN & 0.890 & [0.807, 0.958] & & \\

\midrule

\multirow{2}{*}{2}
 & BISN & 0.976 & [0.952, 0.994]
 & \multirow{2}{*}{0.029\ [$-$0.007, 0.069]}
 & \multirow{2}{*}{0.118} \\
 & DANN & 0.947 & [0.896, 0.986] & & \\

\midrule

\multirow{2}{*}{3}
 & BISN & 0.893 & [0.811, 0.959]
 & \multirow{2}{*}{0.050\ [0.006, 0.110]}
 & \multirow{2}{*}{0.021} \\
 & DANN & 0.843 & [0.717, 0.941] & & \\

\midrule

\multirow{2}{*}{EWM}
 & BISN & 0.934 & [0.905, 0.959]
 & \multirow{2}{*}{0.041\ [0.013, 0.074]}
 & \multirow{2}{*}{0.002} \\
 & DANN & 0.893 & [0.842, 0.937] & & \\

\bottomrule
\end{tabular}
}

\end{table*}

\clearpage
\section{Learned Convolution Kernels}
\label{app:learned_kernels}

We extracted the final convolution kernel from the validation-selected checkpoint for each LOBO fold. Each kernel was compared with the analytic first-derivative Savitzky--Golay initialisation. Because instance normalisation follows the convolution, overall kernel scale is not directly identifiable. We therefore normalised each kernel to unit \(L_2\) norm before comparing its shape.
The Pearson correlations with the SG initialisation were \(0.05\), \(0.09\), and \(0.37\) for folds 1--3, respectively. Thus, the optimised kernels showed fold-specific unit-norm shape differences ($ 1.4, 1.3, 1.1$, respectively) and weak similarity to their initialisation.

\begin{figure*}[htbp]
\centering
\includegraphics[width=\textwidth]{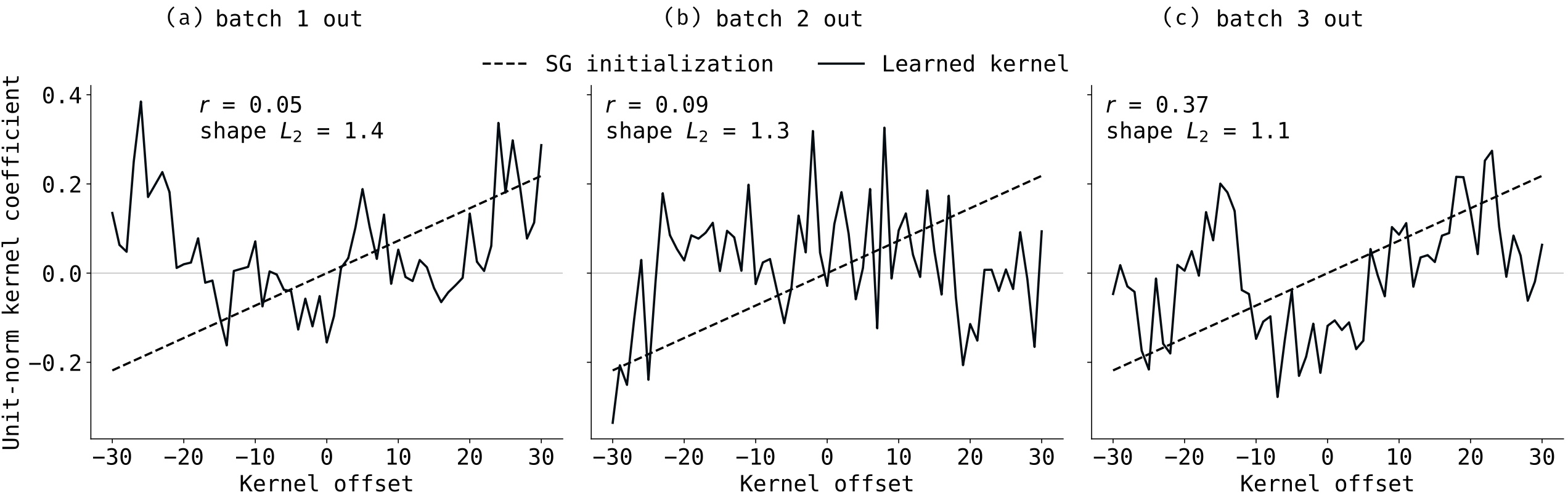}
\caption{Comparison of the analytic first-derivative Savitzky--Golay initialisation and learned BISN convolution kernels across LOBO folds. Kernels were normalised to unit \(L_2\) norm because instance normalisation follows the convolution. The Pearson correlation \(r\) and unit-norm shape difference are reported for each fold (a--c). Individual kernel coefficients were treated as learned signal-processing parameters and were not assigned to specific chemical constituents or absorption processes.}
\label{fig:learned_sg_kernels}
\end{figure*}

\end{document}